\documentclass[a4paper,fleqn]{cas-sc}

\usepackage[authoryear,longnamesfirst]{natbib}

\usepackage{tikz-dependency}
\usepackage{tikz}
\usepackage{tikz-qtree}
\usepackage{latexsym}
\usepackage{amssymb}
\usepackage{amsmath}
\usepackage{arydshln}

\newcommand{\change}[1]{\textcolor{black}{#1}}

\def\tsc#1{\csdef{#1}{\textsc{\lowercase{#1}}\xspace}}
\tsc{WGM}
\tsc{QE}
\begin{document}
\let\WriteBookmarks\relax
\def\floatpagepagefraction{1}
\def\textpagefraction{.001}

\shorttitle{Transition-based Semantic Role Labeling with Pointer Networks}    

\shortauthors{D. Fern\'andez-Gonz\'alez}  

\title [mode = title]{Transition-based Semantic Role Labeling with Pointer Networks}  

%\tnotemark[<tnote number>] 

%\tnotetext[<tnote number>]{<tnote text>} 


\author[1]{Daniel Fern\'{a}ndez-Gonz\'{a}lez}[orcid=0000-0002-6733-2371]

\cormark[1]

%\fnmark[<footnote mark no>]

\ead{d.fgonzalez@udc.es}

\ead[url]{https://danifg.github.io}

\credit{Conceptualization, methodology, software, validation, formal analysis, investigation, data curation, writing - original draft, writing - review \& editing, visualization}
%\credit{<Credit authorship details>}

\affiliation[1]{organization={Universidade da Coru\~{n}a, CITIC, FASTPARSE Lab, LyS Group, Depto. de Ciencias de la Computaci\'{o}n y Tecnolog\'{i}as de la Informaci\'{o}n},
            addressline={Campus de Elvi\~{n}a, s/n }, 
            city={A Coru\~{n}a},
%          citysep={}, % Uncomment if no comma needed between city and postcode
            postcode={15071}, 
            %state={},
            country={Spain}}

% Footnote of the second author

% Email id of the second author



% \affiliation[<aff no>]{organization={},
% %          citysep={}, % Uncomment if no comma needed between city and postcode

\cortext[1]{Corresponding author}



\begin{abstract}
Semantic role labeling (SRL) focuses on recognizing the predicate-argument structure of a sentence and plays a critical role in many natural language processing tasks such as machine translation and question answering. Practically all available methods do not perform full SRL, since they rely on pre-identified predicates, and most of them follow a pipeline strategy, using specific models for undertaking one or several SRL subtasks. In addition, previous approaches have a strong dependence on syntactic information to achieve state-of-the-art performance, despite being syntactic trees equally hard to produce. These simplifications and requirements make the majority of SRL systems impractical for real-world applications. In this article, we propose the first transition-based SRL approach that is capable of completely processing an input sentence in a single left-to-right pass, with neither leveraging syntactic information nor resorting to additional modules. Thanks to our implementation based on Pointer Networks, full SRL can be accurately and efficiently done in $O(n^2)$, achieving the best performance to date on the majority of languages from the CoNLL-2009 shared task.

\end{abstract}



% Keywords

\maketitle

% Main text
\section{Introduction}
\begin{figure}
\centering
\small
\includegraphics[width=0.6\columnwidth]{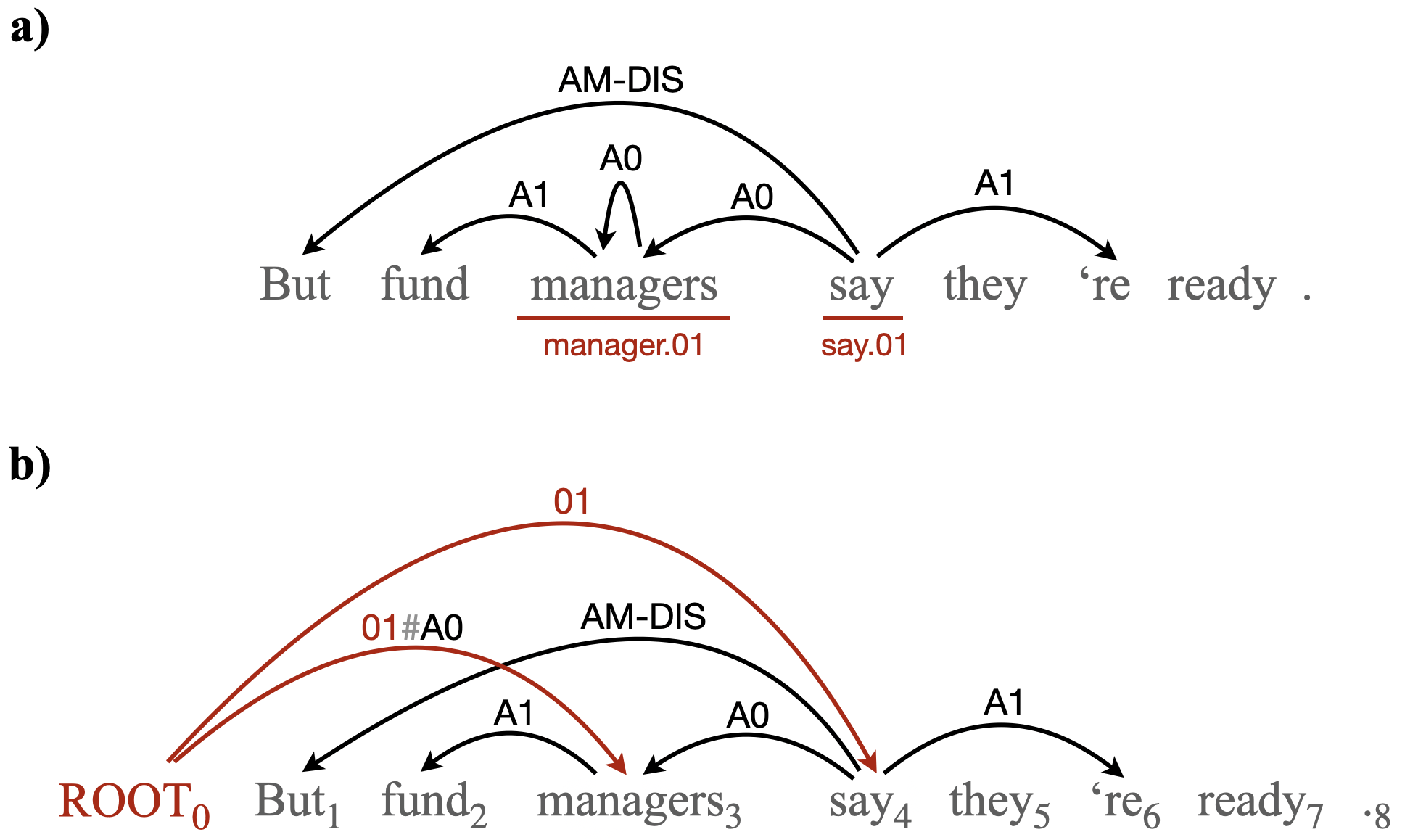}
\caption{\textbf{a)} Example of a dependency-based SRL structure from the CoNLL-2009 English data and \textbf{b)} its labeled dependency graph representation for end-to-end modeling.}
\label{fig:example}
\end{figure}

\textit{Semantic role labeling} (SRL) has been successfully applied to a wide spectrum of \textit{natural language processing} (NLP) applications such as \textit{machine translation} \citep{shi-etal-2016-knowledge,WangIJCAI2016,marcheggiani-etal-2018-exploiting}, \textit{information extraction} 
 \citep{bastianelli-etal-2013-textual},  \textit{question answering}  \citep{yih-etal-2016-value,zheng-kordjamshidi-2020-srlgrn,xu-etal-2020-semantic}
 and \textit{text comprehension} \citep{zhang2019explicit}, \textit{inter alia}.
 This fundamental NLP task can be seen as a shallow semantic parsing that aims to extract the \textit{``who did what to whom, how, where and when''} from an input text by identifying \textit{predicate-argument} relations. These semantic relations are usually represented by a set of 
 %predicate-argument 
 labeled dependencies, where each one connects a predicate to either the entire phrasal argument (following the \textit{span-based} SRL formalism) or just the argument's syntactic head (following the \textit{dependency-based} SRL annotation). \change{While we can find recent studies that seek to improve performance on the former \citep{he-etal-2018-jointly,strubell-etal-2018-linguistically,zhang-etal-2022-semantic},} this research work focuses on 
 %the latter, 
 dependency-based SRL,
 which was popularized by CoNLL-2008 and CoNLL-2009 shared tasks \citep{surdeanu-etal-2008-conll,hajic-etal-2009-conll}. An example of predicate-argument relations represented as a dependency-based SRL structure is depicted in Figure~\ref{fig:example}(a).

 SRL is traditionally decomposed into four simpler subtasks: \textit{predicate identification} (\textit{e.g.}, \textit{managers} in Figure~\ref{fig:example}(a)), \textit{predicate sense disambiguation} \change{(\textit{manager.01} is the sense of predicate \textit{managers} in the example)}, \textit{argument identification} (\textit{e.g., fund}) and \textit{argument role labeling} (\textit{fund} is argument \texttt{A1} for predicate \textit{managers}). While the four subtasks had to be completed in the CoNLL-2008 shared task, CoNLL-2009 corpora notably simplified SRL by providing pre-identified predicates beforehand. This simplification, coupled with the fact that the vast majority of approaches are only tested on the CoNLL-2009 benchmark, resulted in the common practice of not performing full SRL and exclusively focusing on the last three subtasks. 
 
 %Furthermore, we can find in the literature as most SRL systems adopt a \textit{pipeline} framework, resorting to different specific models to separately address one or two subtasks \citep{roth-lapata-2016-neural,marcheggiani-etal-2017-simple,he-etal-2018-syntax,cai-lapata-2019-syntax}; and some recent studies that follow an \textit{end-to-end} strategy, employing a single model to accomplish the SRL task. \citep{he-etal-2019-syntax,li-etal-2020-high,conia-navigli-2020-bridging}. However, the wide majority of these end-to-end approaches do not perform predicate identification and, therefore, are not considered full SRL systems.
 
 \change{Furthermore, we can find in the literature as most SRL systems adopt either a \textit{pipeline} framework \citep{roth-lapata-2016-neural,marcheggiani-etal-2017-simple,he-etal-2018-syntax,cai-lapata-2019-syntax} or an \textit{end-to-end} strategy \citep{he-etal-2019-syntax,li-etal-2020-high,conia-navigli-2020-bridging}. While the former approach resorts to different specific models to separately address one or two subtasks, the latter strategy employs a single model to accomplish the SRL task. However, it is worth mentioning that the vast majority of these end-to-end approaches do not perform predicate identification and, therefore, are not considered full SRL systems.}

 \change{Moreover, most previous efforts 
 mainly focused on \textit{syntax-aware} SRL methods \citep{he-etal-2018-syntax,he-etal-2019-syntax,cai-lapata-2019-syntax,kasai-etal-2019-syntax,li-etal-2021-syntax}: \textit{i.e.}, they leverage syntactic information (also provided by the CoNLL-2009 corpora) to produce state-of-the-art accuracies. These approaches were motivated, among other reasons, by the fact that a significant portion of arcs from syntactic dependency trees matches predicate-argument relations in 
 %dependency-based 
 SRL structures.} Nevertheless, it is important to note that dependency parsing is also an equally challenging and resource-consuming NLP task, and syntactic training data can be especially scarce in low-resource languages.
 
 The fact that practically all 
 %proposed 
 approaches heavily rely on information that is not always available (such as gold predicates and syntactic dependency trees) makes them impractical for real-life downstream applications that require full SRL. In addition, the use of a single model in end-to-end architectures not only mitigates the error-propagation problem 
 %between modules 
 in pipeline strategies, but also notably simplifies the decoding process. %, being especially appealing for those scenarios where decoding speed is crucial.
 
 To alleviate these inconveniences, 
 %In order to propose a practical SRL system, 
 \citet{cai-etal-2018-full} introduced the first end-to-end \textit{syntax-agnostic} approach for accurately performing full SRL. \change{They handle the four subtasks 
 %(including predicate identification) 
 by a single \textit{graph-based} model}, following a two-stage decoding procedure: they first identify and classify all predicates and then search for arguments and semantic roles for each of these predicates. \change{The latter is implemented by independently scoring all possible predicate-argument dependencies and then exhaustively searching for a
high-scoring graph
by combining these scores.} 
 %They address SRL as a graph parsing task uniformly handling the four SRL subtasks (including predicate identification) by a single model, without requiring any additional syntactic information. To achieve that, they apply the widely-used biaffine attention mechanism \citep{DozatM17} 
 %for graph-based dependency parsing, they  
 %for exhaustively scoring all possible predicate-argument dependencies and labels, and, during decoding, they first identify and classify all predicates and, in a second stage, they identify arguments and semantic roles for each of these predicates. 
%OLD
%  This work was recently improved by \citet{zhou2021fast} that 
%   proposed a syntax-agnostic graph-based method that jointly identifies and classifies predicates and arguments in a single stage, obtaining a remarkable performance in full end-to-end SRL. 
\change{This work was recently improved by \citet{zhou2021fast}. They 
proposed a syntax-agnostic method that jointly identifies and classifies predicates and arguments in a single stage. In addition, their approach leverages high-order information, scoring sets of predicate-argument dependencies and computing the high-scoring graph in cubic time. The resulting graph-based model achieves a remarkable performance in full end-to-end SRL.}

 Unlike graph-based methods, \textit{transition-based} algorithms were barely proposed for
 SRL modeling. \change{These generate a sequence of actions (transitions) to incrementally build predicate-argument dependencies (usually from left to right). This is typically done by local, greedy prediction and can efficiently process a sentence 
in a linear or quadratic number of actions.}
 %These incrementally build predicate-argument dependencies by applying a sequence of transitions. 
 \change{Although they provide higher efficiency than graph-based models} and were successfully employed in other parsing tasks \citep{ma-etal-2018-stack,fernandez-gonzalez-gomez-rodriguez-2019-left,fernandez-gonzalez-gomez-rodriguez-2020-transition,FerGomKBS2022}, only two transition-based SRL systems were presented \citep{choi-palmer-2011-transition,fei-end-SRL}, neither of them performing full syntax-agnostic SRL.

In this article, we propose the first transition-based approach for full syntax-agnostic SRL.\footnote{Source code available at \url{https://github.com/danifg/SRLPointer}.}
%\footnote{\change{Source code will be publicly available after acceptance.}} 
Our model does not rely on any kind of syntactic information, even discarding part-of-speech (PoS) tags,\footnote{PoS tags are considered lexical-level syntactic features and, therefore, a truly syntax-agnostic approach should not leverage that information.} and 
%jointly identifies and processes predicates and arguments, 
%as well as the relation between them, 
incrementally produces predicate-argument dependencies
%a dependency-based SRL structure 
in a single left-to-right pass. For implementing our technique, we resort to Pointer Networks \citep{Vinyals15}, which provide an efficient $O(n^2)$ runtime complexity in practice. We experimentally prove that our end-to-end SRL system surpasses strong baselines (including syntax-aware approaches) on the CoNLL-2009 corpora, becoming the
%as well as it is on par with 
highest-performing model in practically all languages. Our major contributions can be summarized as follows:
\begin{itemize}
    \item We design the first syntax-agnostic approach for transition-based SRL, requiring neither syntactic dependency trees nor PoS tag information.
    \item Our approach is end-to-end and can be directly applied to plain text, tackling the four SRL subtasks in one shot and without requiring any external module.
    \item Our model is robust and achieves state-of-the-art results on the majority of languages from
    CoNLL-2009 benchmark with and without pre-identified predicates.
    \item We empirically prove that the proposed transition-based technique processes CoNLL-2009 corpora in $O(n^2)$ time, being more efficient than best-performing graph-based models ($O(n^3)$).
    %\item Source code will be publicly available after acceptance.
\end{itemize}
 
 %surpassing strong baselines and 

The remainder of this article is organized as follows:
Section~\ref{sec:related} introduces previous studies for 
%end-to-end 
syntax-agnostic SRL. In Section~\ref{sec:model}, we 
%first explain how end-to-end dependency-based SRL can be undertaken as a graph parsing task; secondly, we
%Section~\ref{sec:transitions} 
define the transition system for full SRL and
%, finally, we 
%used in our approach. In Section~\ref{sec:arch}, we 
detail the proposed Pointer Network architecture. In Section~\ref{sec:experiments}, we extensively evaluate our SRL model on CoNLL-2009 corpora, present a discussion of the experimental results, analyze the contribution of each component and study the time complexity of our approach in practice.  
Lastly, Section~\ref{sec:conclusion} includes final conclusions.

%TODO: Añadir papers más nuevos en syntax-aware, pero son demasiados habría que poner inter alia o poner los más recientes.
\section{Related work}
\label{sec:related}
Syntax-based approaches %\citep{roth-lapata-2016-neural,marcheggiani-etal-2017-simple,he-etal-2018-syntax,he-etal-2019-syntax,cai-lapata-2019-syntax,kasai-etal-2019-syntax}
%fei-etal-2021-better
\citep{roth-lapata-2016-neural,he-etal-2018-syntax,he-etal-2019-syntax,cai-lapata-2019-syntax,li-etal-2021-syntax} have been the mainstream 
%since CoNLL-2008 and CoNLL-2009 shared tasks to date
for dependency-based SRL, consistently proving that syntactic information is highly effective for achieving state-of-the-art accuracies. 

%However, syntactic information is not only scarce in low-resource settings, but also equally hard to generate.

Alternatively, \citet{marcheggiani-etal-2017-simple} proposed the first syntax-agnostic 
%graph-based 
model for dependency-based SRL. Instead of leveraging syntactic features for capturing long-distance predicate-argument dependencies, they employ a BiLSTM-based encoder. 
%This produces an encoding of the whole sentence for each specific predicate that will be used for predicting its arguments. 
Their approach exclusively addresses argument identification and labeling, directly using gold predicates from CoNLL-2009 corpora and resorting to other works \citep{zhao-etal-2009-multilingual-dependency,bjorkelund-etal-2009-multilingual,roth-lapata-2016-neural} for language-specific predicate disambiguation. This initial attempt was followed by the graph-based model by \citet{cai-etal-2018-full}, which is considered the first syntax-agnostic method for full end-to-end SRL. \change{They 
perform the four SRL subtasks by a single model.} 
%(including predicate identification and classification). 
To achieve that, they apply the widely-used \textit{biaffine} attention mechanism \citep{DozatM17} 
 for exhaustively scoring all predicate-argument relations and their semantic roles; then, during decoding, they first search for predicates and, in a second stage, each predicate is processed by identifying and labeling its arguments. %Unlike the work by \citet{marcheggiani-etal-2017-simple}, they use a one-pass BiLSTM-based encoder, producing a single sentence representation for the whole process. 
 This work was improved by other syntax-agnostic graph-based techniques such as 
 %the first-order approach by 
 \citep{li2019dependency} and 
 %the second-order method by 
 \citep{li-etal-2020-high} that, instead of applying a two-stage strategy, model dependency-based SRL as a graph parsing task, where predicates and arguments are uniformly treated and jointly processed. However, while performing full SRL, these two systems follow a pipeline strategy and, therefore, are not considered end-to-end: the former follows \citep{roth-lapata-2016-neural} for predicate disambiguation and the latter identifies predicates in advance with a separate sequence tagging model.  Recently, \citet{zhou2021fast} extended the work by \citet{li-etal-2020-high} to full end-to-end SRL, obtaining promising results on the CoNLL-2009 English dataset without pre-identified predicates. 
 
 We can also find recent syntax-agnostic 
 %graph-based 
 approaches that do not perform full SRL and follow a predicate-centered strategy for decoding and word representation (based on gold predicates provided by CoNLL-2009 corpora): \citep{chen-etal-2019-capturing} and \citep{lyu-etal-2019-semantic}, which additionally
 implement different iterative refinement procedures, and \citep{conia-navigli-2020-bridging}, which employs an additional specific encoder for contextualizing each gold predicate in the sentence. 
 %These studies do not perform full SRL and follow a predicate-centered strategy for the neural architecture development, strongly relying on pre-identified predicates and handling them differently.

Regarding transition-based SRL modeling, just two syntax-aware attempts were proposed. \citet{choi-palmer-2011-transition} presented a pre-deep-learning system that relies on handcrafted syntactic features and is only able to identify arguments and predict their semantic roles. And, recently, \citet{fei-end-SRL} developed an
%full 
end-to-end system that resorts to TreeLSTMs \citep{tai-etal-2015-improved} for leveraging syntactic information. In addition, this work firstly processes the sentence from left to right in order to identify any possible predicate and, when a predicate is found, their transition system searches for arguments from near to far. Finally, while their approach can be used for full SRL, \citet{fei-end-SRL} did not evaluate their transition-based model on CoNLL-2009 corpora without gold predicates, just testing it on the English dataset with pre-identified predicates.

On the other hand, graph-based approaches \citep{wang-etal-2019-second,bertbaseline} are also the mainstream in other semantic parsing tasks such as Semantic Dependency Parsing (SDP) \citep{oepen-etal-2014-semeval}; however,
\citet{fernandez-gonzalez-gomez-rodriguez-2020-transition} introduced a transition-based algorithm that yields state-of-the-art scores on that task. Inspired by the latter,
we design the first transition-based model that, without any kind of syntactic features, performs full end-to-end SRL in a single 
forward pass.

\section{Model}
\label{sec:model}
\subsection{Graph representation}
\label{sec:sdp}
%For full end-to-end SRL modeling, the four subtasks must be formulated as a single semantic parsing task, where predicates and arguments are uniformly treated and the relations between them are represented as labeled dependency arcs, being the predicate the semantic \textit{head}, the argument the semantic \textit{dependent} and the semantic role the \textit{dependency label}. 
\change{For full end-to-end SRL modeling, the four subtasks must be formulated as a single %semantic 
graph parsing task, where predicates and arguments are uniformly treated. To that end, the relations between predicates and arguments must be represented as labeled dependency arcs, where the predicate acts as the semantic \textit{head}, the argument as the semantic \textit{dependent} and the semantic role as the \textit{dependency label}.} In addition, end-to-end graph-based SRL models \citep{cai-etal-2018-full,he-etal-2018-syntax,li2019dependency} augment the original dependency-based structure by building a single-rooted graph (not necessarily acyclic): \textit{i.e.},
all predicates are attached to an artificial root node (added at the begging of the sentence) and the resulting arcs are labeled with 
predicate sense tags.\footnote{\change{As common practice, the lemma is removed from the predicate sense tag (\textit{e.g.,} \textit{manager} in \textit{manager.01}), since that information is typically provided in datasets from CoNLL-2009.}} %and can be easily recovered)}.
Formally, given an input sentence $X = w_0, w_1, \dots, w_n$ (being $w_0$ the artificial root node), a
full 
end-to-end SRL system is expected to completely produce a graph $G$ represented as a set of labeled predicate-argument relations: $G \subseteq W \times W \times L$, where $W$ is the set of input words ($W = \{w_0, w_1, \dots, w_n \}$) and $L$ refers to the set of semantic role plus predicate sense labels. We adopt this graph representation for implementing our transition-based model. \change{In Figure~\ref{fig:example}(b), we present the resulting single-rooted graph obtained from the original dependency-based SRL structure in Figure~\ref{fig:example}(a).} 

\change{Moreover, it is worth noting that, in some languages such as Czech, the same word can serve as two or more arguments to the same predicate, resulting in two or more dependencies between the same two words.
%there can be more than one dependency from a predicate to the same word
%the same predicate and argument 
%\change{(\textit{i.e.}, the same token serves as two or more arguments to the same predicate)}. 
For instance, we can find two dependency arcs between predicate $w_p$ 
%\textit{subjekty} 
and word $w_a$ respectively tagged % the same argument 
%\textit{vztahy} 
with semantic roles \texttt{A1} and \texttt{A2} (meaning that $w_a$ serves as arguments \texttt{A1} and \texttt{A2} for predicate $w_p$). We handle this by keeping just one dependency between words $w_p$ and $w_a$, and by assigning the concatenation of both semantic roles (\texttt{A1|A2}) as dependency label. Additionally, in some datasets, a predicate can be also an argument of itself (\textit{i.e.}, a dependency where the head and the dependent are the same word). An example of this can be seen in Figure~\ref{fig:example}, where the word \textit{managers} acts as the predicate and argument \texttt{A0}. %of the dependency arc labeled with the semantic role \texttt{A0}. 
To encode that information in our final representation, we concatenate the semantic role label of this dependency (\texttt{A0} in the example) with the predicate sense (\texttt{01} for predicate \textit{managers})
and use 
%it
the resulting label (\texttt{01\#A0})
for tagging the arc that will attach that predicate to the artificial root. In both
%cases,
pre-processing strategies,
the original structure is easily recovered before evaluation. }

\subsection{Transition system}
 \label{sec:transitions}
Inspired by \citep{fernandez-gonzalez-gomez-rodriguez-2020-transition}, we design a transition system for generating a graph $G$ for the input sentence by applying a sequence of actions  $A = {a_1, \dots, a_t}$. \change{These actions (or  \textit{transitions}) will be sequentially predicted by a neural model.} %described in the following section.}
%Starting from an initial \textit{parsing configuration}, this algorithm applies a sequence of actions, traversing different configurations until reach a terminal one are state machines
\change{In this section, we formally define %below 
%its main components: 
the main components of the proposed transition system: \textit{state configurations} and 
%available 
\textit{actions}.} %(or  \textit{transitions}).

While other transition-based SRL systems require more complex state configurations with several stacks and additional data structures for temporarily storing partially-processed words \citep{choi-palmer-2011-transition,fei-end-SRL}, we just need to implement two pointers for building any dependency graph.
%on the current focus word. 
More in detail, the proposed transition system has state configurations of the form $c = \langle i, j, \Sigma \rangle$, where $i$ points at the word $w_i$ currently being processed, $j$ indicates the position of the last identified predicate $w_j$ for $w_i$
%for current focus word $w_i$ 
and $\Sigma$ contains the set of already-created edges. Given a sentence $X = w_0, w_1, \dots, w_n$ (with $w_0$ as artificial root node), the process starts at the initial state configuration $c_{initial} = \langle 1, -1,\emptyset \rangle$, where $i$ is pointing at the first input word $w_1$, no predicate position has been saved yet at $j$ and $\Sigma$ is empty. Then, after applying a sequence of actions $A$, the transition system reaches a final configuration of the form $c_{terminal} = \langle n+1, -1, \Sigma \rangle$, where all the words have been shifted 
%and $i$ points at the end of the sentence, 
and $\Sigma$ contains the edges of the graph $G$ for the input sentence $X$.

Unlike the works by \citet{choi-palmer-2011-transition} and \citet{fei-end-SRL} that define six different actions to produce SRL structures, we just require two transitions: 
\begin{itemize}
    \item \textsc{Arc-}\textit{p} that attaches the current focus word $w_i$ to the head word at position $p$, building a semantic dependency arc from the identified predicate $w_p$ to 
    %one of its 
    argument $w_i$. 
    %Formally, 
    By applying this action, the transition system moves from state configurations $\langle i, j, \Sigma \rangle$ to $\langle i, p, \Sigma \cup \{w_p \rightarrow w_i\} \rangle$. This transition can only be applied if the resulting edge has not been created yet (\textit{i.e.}, $w_p \rightarrow w_i \notin \Sigma$) and the predicate $w_p$ is in a higher position than the last identified predicate for $w_i$ in position $j$ (\textit{i.e.}, $j < p$). The latter condition is necessary since the head assignment to the current focus word must follow the left-to-right order used for training.\footnote{Please note that, while applying a different order in head attachments (\textit{e.g.}, inside-out \citep{ma-etal-2018-stack}) could lead to slight improvements in accuracy, 
    we decided to keep it simple and follow a left-to-right strategy.} Finally, the resulting dependency arc is labeled by a jointly-trained classifier, as described in the following section.
    %how dependency arcs created by the \textsc{Arc-}\textit{p} transition are labelled for producing a well-formed SRL structure.
     \item \textsc{Shift} that moves $i$ one position to the right, pointing at the word $w_{i+1}$, and, since we will start searching for predicates for that unprocessed word, $j$ is initialized to $-1$. Therefore, we move from state configurations $\langle i, j, \Sigma \rangle$ to $\langle i+1, -1, \Sigma \rangle$ by using this action.
\end{itemize} 
\noindent The resulting transition system processes an input sentence from left to right by applying a sequence of \textsc{Shift-Arc} actions that attaches some words to one or several predicates and leaves others unattached, incrementally building a dependency graph. Please see in Table~\ref{tab:transitions} how this transition-based algorithm generates the graph in Figure~\ref{fig:example}(b).

\begin{table}
\begin{center}
\small
\begin{tabular}{@{\hskip 1.0pt}lcccc@{\hskip 1.0pt}}
\hline\noalign{\smallskip}
transition & state configuration & focus word$_i$ & last predicate$_j$  & added arc  \\
\noalign{\smallskip}\hline\noalign{\smallskip}
 & $\langle 1, -1, \Sigma \rangle$ & But$_1$ & &  \\
\textsc{Arc-}\textit{4} & $\langle 1, 4, \Sigma \cup \{4 \rightarrow 1\} \rangle$ & But$_1$ & say$_4$ & say$_4$ $\rightarrow$ But$_1$ \\
\textsc{Shift} & $\langle 2, -1, \Sigma \rangle$ & fund$_2$ &   & \\
\textsc{Arc-}\textit{3} & $\langle 2, 3, \Sigma \cup \{3 \rightarrow 2\} \rangle$ & fund$_2$ & managers$_3$  & managers$_3$ $\rightarrow$ fund$_2$ \\
\textsc{Shift} & $\langle 3, -1, \Sigma \rangle$ & managers$_3$ &   & \\
\textsc{Arc-}\textit{0} & $\langle 3, 0, \Sigma \cup \{0 \rightarrow 3\} \rangle$ & managers$_3$ & \textsc{Root}$_0$  & \textsc{Root}$_0$ $\rightarrow$ managers$_3$ \\
\textsc{Arc-}\textit{4} & $\langle 3, 4, \Sigma \cup \{4 \rightarrow 3\} \rangle$ & managers$_3$ & say$_4$  & say$_4$ $\rightarrow$ managers$_3$ \\
\textsc{Shift} & $\langle 4, -1, \Sigma \rangle$ & say$_4$ &   & \\
\textsc{Arc-}\textit{0} & $\langle 4, 0, \Sigma \cup \{0 \rightarrow 4\} \rangle$ & say$_4$ & \textsc{Root}$_0$  & \textsc{Root}$_0$ $\rightarrow$ say$_4$ \\
\textsc{Shift} & $\langle 5, -1, \Sigma \rangle$ & they$_5$ &   & \\
\textsc{Shift} & $\langle 6, -1, \Sigma \rangle$ & 're$_6$ &   & \\
\textsc{Arc-}\textit{4} & $\langle 6, 4, \Sigma \cup \{4 \rightarrow 6\} \rangle$ & 're$_6$ & say$_4$  & say$_4$ $\rightarrow$ 're$_6$ \\
\textsc{Shift} & $\langle 7, -1, \Sigma \rangle$ & ready$_7$ &   & \\
\textsc{Shift} & $\langle 8, -1, \Sigma \rangle$ & .$_8$ &   & \\
\textsc{Shift} & $\langle 9, -1, \Sigma \rangle$ & &   & \\
\noalign{\smallskip}\hline
\end{tabular}
\caption{Transition sequence 
%(first column) 
and resulting state configurations
%pointers $i$ and $j$
for incrementally generating arcs of the dependency graph in Figure~\ref{fig:example}(b).} \label{tab:transitions}
\end{center}
\end{table}

While the described transition system was originally designed for 
%completely 
performing full end-to-end SRL, it can be easily adapted to leverage gold predicate information provided by the CoNLL-2009 corpora (where no predicate identification is required). For that purpose, a third condition must be added to the \textsc{Arc-}\textit{p} transition: it can be applied only if the word $w_p$ is a gold predicate. In addition, each word $w_i$ that is a gold predicate is directly attached to the artificial root $w_0$. \change{Please note that, while these modifications allow a fairer comparison on the CoNLL-2009 benchmark, our approach does not follow a predicate-centered strategy as the vast majority of SRL systems evaluated on that shared task. These typically base their training and decoding procedures on the existence of pre-identified predicates and, as a consequence, they simply focus on individually processing each given predicate by searching for its arguments over the whole input.}

%Finally, we will see in Section~\ref{sec:complexity} as the proposed transition system processes an input sentence in a linear number of actions in practice.
%as described in the following section.

%\subsection{Neural architecture}
\subsection{Pointer Network}
 \label{sec:arch}
We use a Pointer Network \citep{Vinyals15} for implementing the proposed transition-based algorithm. We manage to properly represent state configurations in this neural model and use that information for generating transition sequences necessary for processing input sentences.
%These neural models employ an attention mechanism \citep{Bahdanau2015NeuralMT,luong-etal-2015-effective} to select, at each decoding step, a position from the input. We will use that information to output one of the two available transitions. 
We detail below the different components of our neural architecture:

\paragraph{Word representation} Each input token $w_i$ is represented by the concatenation of word ($e^{word}_i$), lemma ($e^{lemma}_i$) and character-level ($e^{char}_i$) embeddings. The latter is obtained by encoding characters inside $w_i$ with convolutional neural networks (CNNs)
\citep{ma-hovy-2016-end} and, unlike most SRL systems, we do not include PoS tag embeddings in order to develop a truly syntax-agnostic approach. In addition, we also evaluate our model with deep contextualized word embeddings ($e^{BERT}_i$) from pre-trained language model BERT \citep{devlin-etal-2019-bert}. In particular, we use mean pooling (\textit{i.e.}, the average value of all subword embeddings) to extract word-level representations from weights of 
%one or several 
the second-to-last layer. Lastly, we follow previous works \citep{cai-etal-2018-full,he-etal-2018-syntax,li-etal-2020-high} and leverage a predicate indicator embedding ($e^{indicator}_i$)\footnote{This embedding simply marks whether $w_i$ is a predicate or not based on the information provided by CoNLL-2009 corpora.} when our model is tested following the CoNLL-2009 setting and, therefore, pre-identified predicates are available. When all these embeddings are exploited, the word representation $e_i$ is obtained as follows:
$$e_i = e^{word}_i \oplus e^{lemma}_i \oplus e^{char}_i \oplus e^{BERT}_i \oplus e^{indicator}_i$$

%REVISOR1 no estaba de acuerdo
%Finally, it can be argued that, since it has been proved that pre-trained language models encode syntactic information \citep{hewitt-manning-2019-structural}, leveraging deep contextualized word embeddings in syntax-agnostic SRL systems would turn them into syntax-aware approaches. Therefore, strictly speaking, a truly syntax-agnostic SRL system should leverage neither PoS tag nor deep contextualized word embeddings.

\paragraph{Encoder} As a common practice, we feed a three-layer bidirectional LSTM (BiLSTM) \citep{LSTM} to generate a context-aware vector representation $h_i$ for each word vector $e_i$: 
$$ h_i  = \mathrm{BiLSTM}(e_i) = f_i \oplus b_i$$
where $f_i$ and $b_i$ are respectively the forward and backward hidden states of the last LSTM layer at the $i$th position. Additionally, a randomly-initialized  vector $h_0$ is 
used
%prepended at the beginning of the resulting sequence of encoder hidden states 
for denoting the artificial root node. As a result, the sequence of word representations $E = e_1, \dots, e_n$ is encoded into a sequence of encoder vectors $H = h_0, h_1, \dots, h_n$.

\paragraph{Decoder} We employ a unidirectional one-layer
LSTM plus an attention mechanism for decoding. Firstly, %being $w_i$ the current focus word and $w_j$ its last assigned predicate (if available),
%pointed by $i$, 
at each time step $t$, state configurations $c_t=\langle i, j, \Sigma \rangle$ are encoded by feeding the LSTM with the combination\footnote{Instead of concatenating both vectors, we compute the element-wise summation to avoid increasing the dimension of the resulting vector $r_t$.} of the respective encoder representations $h_i$ and $h_j$ of the current focus word ($w_i$) and its last assigned predicate ($w_j$) if available. This will generate the state configuration representation $s_t$:\footnote{\change{Please note that $\Sigma$ in $c_t$ is not 
%directly 
used for generating the state configuration representation $s_t$. $\Sigma$ was exclusively designed to collect already-built edges and its main purpose is to prevent the transition system from creating dependency arcs already added to $\Sigma$.}} 
%ESTO NO ES VERDAD
% However, dependency arcs created in previous state configurations $c_{<t}$ (represented by $r_{<t}$) %(which some of them created dependency arcs included in $\Sigma$) 
% are 
% indeed 
% encoded in the LSTM-based decoder and, therefore, affect the current state configuration representation $s_t$.}
%at time step $t$:
$$r_t = h_{i} + h_{j}$$ 
$$s_t = \mathrm{LSTM}(r_t)$$
Please note that, while introducing high-order information (such as the co-parent representation $h_j$ of a future predicate of $w_i$) significantly penalizes graph-based models' runtime complexity,
%in one order of magnitude, 
transition-based algorithms can straightforwardly leverage this information without harming their efficiency.

\begin{figure}%[t]
\centering
\includegraphics[width=\textwidth]{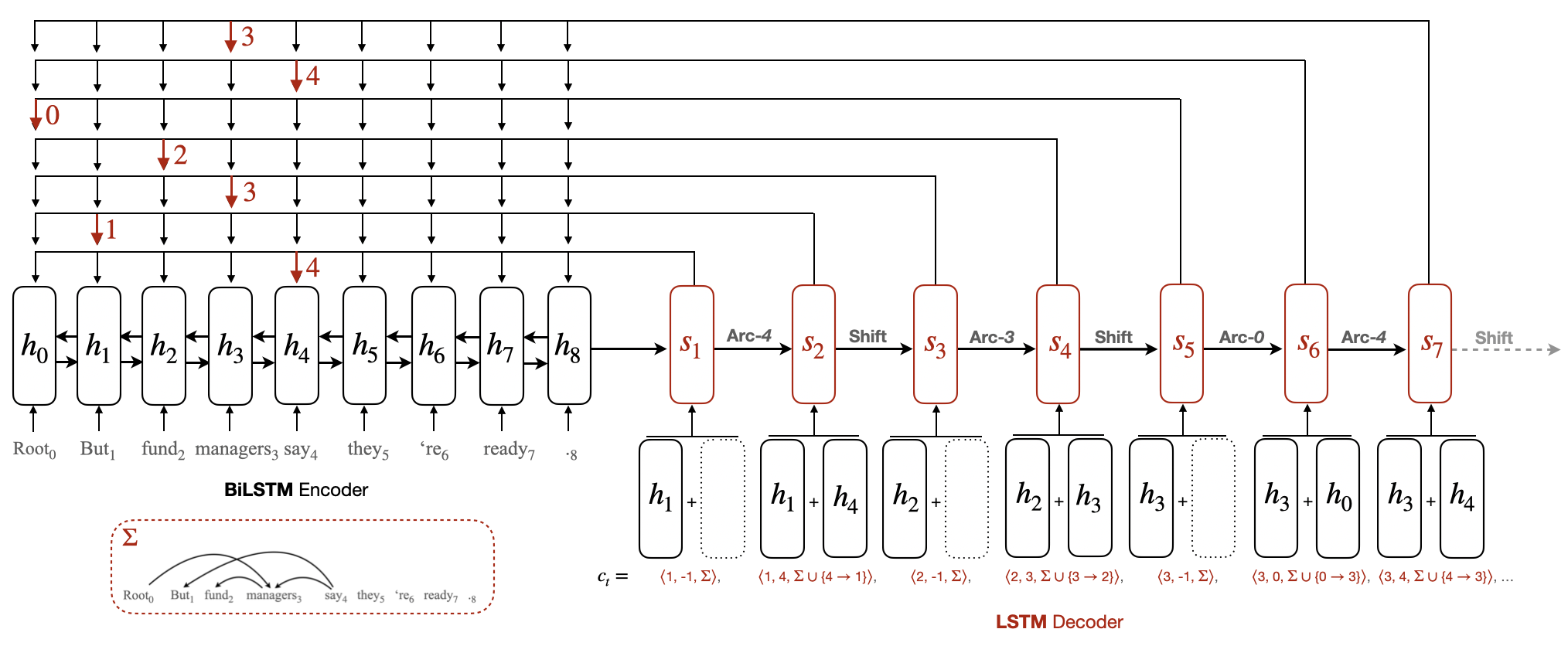}
\caption{Pointer Network architecture and decoding procedure for partially generating the dependency graph in Figure~\ref{fig:example}(b).}
\label{fig:network}
\end{figure}

Once the current state configuration $c_t$ is properly represented as $s_t$, an attention mechanism is used for selecting the action $a_t$ to be applied at time step $t$.
%on the current state configuration $c_t$. 
%The output of this mechanism is the attention vector $\alpha_t$: ; and
%This mechanism is implemented by 
%the attention vector $\alpha_t$ that is an output distribution over the input positions.
In particular, this mechanism employs the biaffine scoring function \citep{DozatM17} to compute the score between each input word $w_k$ (represented by the encoder vector $h_k$ with $k \in [0,n]$) and the state configuration representation $s_t$; and then normalizes the resulting vector $v^t$ of length $n$ to output the attention vector $\alpha_t$, which is a \textit{softmax} distribution with dictionary size equal to the length of the input:
%\begin{align*}
$$v^t_k = \mathrm{score}(s_t, h_k)= f_1(s_t)^T W f_2(h_k)\\
+U^Tf_1(s_t) + V^Tf_2(h_k) + b$$
$$\alpha_t = \mathrm{softmax}(v^t)$$
%\end{align*}
where $W$ is the weight matrix of the bi-linear term, $U$ and $V$ are the weight tensors of the linear terms, $b$ is the bias vector and 
$f_1(\cdot)$ and $f_2(\cdot)$ are two one-layer perceptrons with ELU activation to obtain lower-dimensionality \change{and avoid overfitting}.  From the attention vector $\alpha_t$, we select the highest-scoring position $p_t$ from the input
%$$p_t = \underset{k}{\arg \max}(\alpha_t, k)$$
and, being  $c_t=\langle i, j, \Sigma \rangle$, use that information to choose the current action $a_t$ between the two available transitions as follows:
\begin{itemize}
    \item if $p_t=i$, then a \textsc{Shift} action is applied, moving the focus word pointer to the next word.
    \item On the contrary, if $p_t \neq i$, then the \textsc{Arc} transition parameterized with $p_t$ will be considered, building a dependency arc between the word $w_i$ and its predicate $w_{p_t}$. In case that conditions required by this action are not satisfied, then the next highest-scoring position in $\alpha_t$ will be used for choosing again between the \textsc{Shift} and \textsc{Arc} transitions.
\end{itemize}
In Figure~\ref{fig:network}, we include a sketch of the proposed Pointer Network architecture and decoding steps for partially building the graph structure in Figure~\ref{fig:example}(b).

Finally, 
%In particular, 
when an \textsc{Arc}-$p_t$ transition creates an edge between the current focus word $w_i$ %(encoded as $s_t$) 
and the predicate $w_{p_t}$,
%(represented by $h_{p_t}$), 
we apply a classifier for labeling it. 
%In particular, 
This labeler is implemented by the %previously-described 
biaffine scoring function previously described. \change{Concretely, for each available semantic role or predicate sense label $l \in L$, we 
%first 
%use it for 
compute the score of assigning $l$ to the predicted arc between the argument $w_i$ (encoded as $s_t$) and the predicate $w_{p_t}$ (represented by $h_{p_t}$) as follows:}
$$u^{t}_l = \mathrm{score}(s_t, h_{p_t}, l)= g_1(s_t)^T W_l g_2(h_{p_t})
+U_l^Tg_1(s_t) + V_l^Tg_2(h_{p_t}) + b_l$$
$$\beta_t = \mathrm{softmax}(u^t)$$
%where $g_1(\cdot)$ and $g_2(\cdot)$ are two one-layer perceptrons with ELU activation and $W_l$, $U_l$, $V_l$ and $b_l$ are parameters exclusively used for each label $l$. 
\change{where $W_l$ is a weight matrix, $U_l$ and $V_l$ are weight tensors and $b_l$ is the bias vector exclusively used for each label $l$, and $g_1(\cdot)$ and $g_2(\cdot)$ are two one-layer perceptrons with ELU activation.}
%is a weight matrix, $U_l$ and $V_l$ are weight tensors and bias  is the bias vector are used
Then, we
%for labeling a dependency arc created by the transition system, we compute the score of each label $l \in L$ and
select the highest-scoring label in the vector $\beta_t$ 
%as a candidate 
for tagging the predicted arc. 
%$$l = \underset{k}{\arg \max}(\beta_t, k)$$

\paragraph{Training objectives}
\change{The loss of our model comes from the transition system and the labeler.}
%and both are implemented as cross-entropy loss. 
On the one hand, the transition system is trained by minimizing the total log loss of choosing the correct sequence of \textsc{Shift-Arc} transitions $A$ to output the gold dependency graph $G$ for the input sentence $X$ (\textit{i.e.}, predicting the correct sequence of indices $p_t$, with each decision at time step $t$ being conditioned by previous ones ($p_{<t}$)):
$$\mathcal{L}_{tran}(\theta) = - \sum_{t=1}^{T} log P_\theta (p_t | p_{<t},X)$$

On the other hand, the labeler is trained 
%with softmax cross-entropy 
to minimize the total log loss of assigning the correct label $l$, given a dependency arc 
from predicate $w_{p_t}$ to dependent word $w_i$:
$$\mathcal{L}_{label}(\theta) = - \sum_{t=1}^{T} log P_\theta (l | w_{p_t},w_i)$$

Finally, 
%\textit{We implement being both implemented as cross-entropy loss, and} 
we jointly train the transition system and the labeler by optimizing the sum of their losses: $$\mathcal{L}(\theta) = \mathcal{L}_{tran}(\theta) + \mathcal{L}_{label}(\theta)$$

\section{Experiments}
\label{sec:experiments}
%CL LI 2021. En los experimentos con añadir los que añade Li et al. 2021 y alguno más de 2020 y 2021 (CONIA no creo que lo metamos salvo que le ganemos o mulcaire o FEI 2021 pointernetwork.). Syntax-aware podemos incluir aunque nos ganen.

\begin{table}[tbp]
\centering
\begin{tabular}{@{\hskip 2pt}lcccccc@{\hskip 2pt}}
\toprule
\textbf{Language} & \textbf{Sentences} & \textbf{Annotated} & \textbf{Avg. length} & \textbf{Tokens} & \textbf{Predicates} & \textbf{Arguments}\\
\midrule
Catalan (CA) & 13,200 & 12,873 & 30.2 & 390,302 & 37,431 & 84,367 \\
Chinese (ZH) & 22,277 & 21,071 & 28.5 & 609,060 & 102,813 & 231,869 \\
Czech (CZ) & 38,727 & 38,578 & 16.9 & 652,544 & 414,237 & 365,255 \\
English (EN) & 39,279 & 37,847 & 25.0 & 958,167 & 179,014 & 393,699 \\
German (DE) & 36,020 & 14,282 & 22.2 & 648,677 & 17,400 & 34,276 \\
Spanish (ES) & 14,329 & 13,835 & 30.7 & 427,442 &  43,824 & 99,054  \\
\bottomrule
\end{tabular}
\centering
\setlength{\abovecaptionskip}{4pt}
\caption{Data statistics for CoNLL-2009 training sets. We report the number of sentences, 
%the number of 
annotated sentences (with at least one predicate), 
%the total number of 
tokens, predicates and arguments, as well as the average sentence length.}
\label{tab:stats}
\end{table}

\subsection{Data}
We conduct experiments on datasets from CoNLL-2008 and CoNLL-2009 shared tasks \citep{surdeanu-etal-2008-conll,hajic-etal-2009-conll}. 
%While
The former was an English-only benchmark with in-domain (from the Wall Street Journal corpus \citep{marcus93}) and out-of-domain (from the Brown Corpus \citep{BrownCorpus}) test sets. This was extended by the CoNLL-2009 shared task to a multilingual benchmark by adding 6 more languages (Catalan, Chinese, Czech, German, Japanese and Spanish) with 6 in-domain and 2 out-of-domain test sets. Unlike CoNLL-2008 English dataset, CoNLL-2009 corpora notably simplified the SRL task by providing gold predicates beforehand.\footnote{Please note that the CoNLL-2009 shared task also augmented CoNLL-2008 English dataset with pre-identified predicates.} In order to properly test our model under a real-world usage scenario and also compare it to most previous works, we evaluate our approach on 
%all 
CoNLL-2009 datasets\footnote{We do not evaluate our approach on the Japanese dataset since it is no longer available due to licensing problems.} following two different settings: \textit{w/ pre-identified predicates} and \textit{w/o pre-identified predicates} (requiring the latter full SRL as in the CoNLL-2008 shared task). 

In Table~\ref{tab:stats}, we summarize the training data statistics for each language.
From this information, we can see as, while having a notable amount of sentences in the training set, German is considered a low-resource language 
%in the CoNLL-2009 shared task 
due to the low proportion of annotated sentences and predicate instances in comparison to other languages \citep{conia-navigli-2020-bridging}. For the same reasons, Catalan and Spanish are also classified as low-resource languages. 

Finally, we use the CoNLL-2009 official scoring script\footnote{\url{https://ufal.mff.cuni.cz/conll2009-st/scorer.html}} for performance evaluation. This measures the labeled precision, recall and F$_1$ score for semantic dependencies.

\subsection{Setup}
In our experiments, word and lemma embeddings are initialized with 300-dimensional GloVe vectors \citep{pennington-etal-2014-glove} for English; structured-skipgram embeddings \citep{Ling2015} for Chinese (dimension 80), German (dimension 64) and Spanish (dimension 64); and 64-dimensional Polyglot embeddings for Catalan and Czech. 100-dimensional character-level embeddings are randomly initialized and, for CNNs, we use 100 filters with a window size of 3 and max-pooling. For BERT-based embeddings, we extract respectively 768-dimensional and 1024-dimensional vectors from specific BERT$_{\mathrm{base}}$ and BERT$_{\mathrm{large}}$ models \citep{devlin-etal-2019-bert}:
%dependening on the language being the specific pre-trained versions:
{\tt bert-large-cased} for English, {\tt bert-base-multilingual-cased} for Catalan, {\tt bert-base-chinese} for Chinese, {\tt deepset/gbert-large} \citep{chan-etal-2020-germans} for German, 
%SlavicBert 
{\tt bert-base-bg-cs-pl-ru-cased} \citep{arkhipov-etal-2019-tuning} for Czech and {\tt bert-base-spanish-wwm-cased} \citep{CaneteCFP2020} for Spanish. Following a greener and less resource-consuming strategy, BERT-based embeddings are not fine-tuned during training. Finally, we randomly initialize a 16-dimensional predicate indicator embedding under the \textit{w/ pre-identified predicate} setup.

Most hyperparameters were taken from \citep{fernandez-gonzalez-gomez-rodriguez-2020-transition} and we directly apply them to all datasets and languages without further optimization. For training, we employ Adam optimizer \citep{Adam} with initial learning rate of $\eta_0 =$ 0.001, $\beta_1 =$ 0.9 and $\beta_2=$ 0.9. We also use a fixed decay rate of 0.75 and a gradient clipping of 5.0 in order to mitigate the gradient exploding effect \citep{gradientclipping}. Moreover, we use LSTMs with 512-dimensional hidden states for both encoder and decoder, applying recurrent dropout \citep{recurrent_dropout} with a drop rate of 0.33 between hidden states and layers. We also apply a 0.33 dropout to all embeddings. In addition, all models are trained up to 600 epochs with batch size 32,
and beam-search decoding with beam size 5 is utilized in all experiments. Finally, we choose the checkpoint with the highest labeled F$_1$ score on the development set for posterior in-domain and out-of-domain evaluations.

%All the experiments were performed on an Intel(R) Core(TM) i7-8700K CPU running at 3.70GHz with 32GB of RAM and two GeForce GTX 1080Ti GPUs. 

\begin{table}[tbp]
\centering
\begin{tabular}{@{\hskip 2pt}l@{\hskip 5pt}c@{\hskip 5pt}c@{\hskip 5pt}c@{\hskip 5pt}c@{\hskip 15pt}c@{\hskip 5pt}c@{\hskip 5pt}c@{\hskip 2pt}}
\toprule
& & \multicolumn{3}{l}{\ \ \ \ \ \textbf{WSJ}}
& \multicolumn{3}{c}{\textbf{Brown}}
\\
\textbf{System} & \textbf{\scriptsize{end-to-end}} &\textbf{P} & \textbf{R} & \textbf{F$_1$} & \textbf{P} & \textbf{R} & \textbf{F$_1$} \\
\midrule
\scriptsize{\textit{(Syntax-aware)}}\\
%\citet{zhao-etal-2009-semantic} & n & & & 82.1 & & &  \\
\citet{he-etal-2018-syntax} & n & 83.9 & 82.7 & 83.3 & - & - & - \\
\citet{zhou-etal-2020-parsing} + Joint & n & \textbf{84.2} & \textbf{87.5} & \textbf{85.9} & \textbf{76.5} & \textbf{78.5} & \textbf{77.5} \\
\hdashline[1pt/2pt]
\citet{zhou-etal-2020-parsing} + Joint + BERT$_\mathrm{fine-tuned}$ & n & \textbf{87.4} & \textbf{89.0} & \textbf{88.2} & \textbf{80.3} & \textbf{82.9} & \textbf{81.6} \\
\citet{munir2020} + ELMO & n & 85.8 & 84.4 & 85.1 & 74.6 & 74.8 & 74.7\\
\citet{li-etal-2021-syntax} + ELMO & n & 86.2 & 86.0 & 86.1 & 73.8 & 74.6 & 74.2 \\
\midrule
\scriptsize{\textit{(Syntax-agnostic)}}\\ 
\citet{cai-etal-2018-full} & y & 84.7 & 85.2 & 85.0 & - & - & 72.5\\
\citet{li2019dependency}$^\dagger$ & n & - & - & 85.1 & - & - & - \\
\citet{li-etal-2020-high}$^\dagger$ & n & 86.0 & 85.6 & 85.8 & 74.4 & 73.3 & 73.8\\
\citet{zhou2021fast}$^\dagger$ & y & \textbf{86.7} & 86.2 & 86.5 & \textbf{75.8} & 74.6 & 75.2 \\
\textbf{This work}$^\dagger$ & y & 85.9 & \textbf{88.0} & \textbf{86.9} & 74.4 & \textbf{76.4} & \textbf{75.4} \\
\hdashline[1pt/2pt]
\citet{li2019dependency}$^\dagger$ + ELMO & n & 84.5 & 86.1 & 85.3 & 74.6 & 73.8 & 74.2 \\
\citet{li-etal-2020-high}$^\dagger$ + BERT$_\mathrm{fine-tuned}$ & n & \textbf{88.6} & 88.6 & 88.6 & \textbf{79.9} & 79.9 & 79.9\\
\citet{zhou2021fast}$^\dagger$ + BERT$_\mathrm{fine-tuned}$ & y & 87.6 & \textbf{90.2} & \textbf{88.9} & 79.0 & \textbf{82.2} & \textbf{80.6}\\
\textbf{This work}$^\dagger$ + BERT & y & 87.2 & 89.8 & 88.5 & \textbf{79.9} & 81.8 & 80.4 \\
\bottomrule
\end{tabular}
\centering
\setlength{\abovecaptionskip}{4pt}
\caption{Precision (P), recall (R) and F$_1$ scores obtained by full SRL systems on the CoNLL-2008/CoNLL-2009 English in-domain (Wall Street Journal, WSJ) and out-of-domain (Brown) test sets \textit{w/o pre-identified predicates}. The first block gathers methods enhanced with syntactic information and, the second block, those that are syntax-agnostic. We also indicate in column ``\textit{end-to-end}'' whether approaches follow an end-to-end (y) or a pipeline (n) strategy, using the latter one or more extra models to accomplish full SRL. \textit{+ELMO} and \textit{+BERT} stand for augmentations with deep contextualized word-level embeddings from pre-trained language models ELMO \citep{peters-etal-2018-deep} and BERT, respectively; and \textit{+Joint} means that the system learns SRL jointly with other tasks. 
Please note that we keep BERT-based embeddings frozen in order to avoid increasing the computational cost, and we denote with ``\textit{fine-tuned}'' those systems that do undertake an expensive fine-tuning during training in order to adapt them to SRL. \change{Finally, we mark with $^\dagger$ those truly syntax-agnostic models that do not leverage PoS tag information.}
%Finally, we mark with $^\dagger$ those truly syntax-agnostic models that use neither PoS tag nor deep contextualized word-level embeddings.
}
\label{tab:results1}
\end{table}

%%%%%%%%%%%%%%%%%%%%%%%%%%%%%%%%%%%%%%%%%%%%%%

\begin{table}[tbp]
\centering
\begin{tabular}{@{\hskip 2pt}l@{\hskip 5pt}c@{\hskip 5pt}c@{\hskip 5pt}c@{\hskip 5pt}c@{\hskip 15pt}c@{\hskip 5pt}c@{\hskip 5pt}c@{\hskip 2pt}}
\toprule
& & \multicolumn{3}{l}{\ \ \ \ \ \textbf{WSJ}}
& \multicolumn{3}{c}{\textbf{Brown}}
\\
\textbf{System} & \textbf{\scriptsize{end-to-end}} &\textbf{P} & \textbf{R} & \textbf{F$_1$} & \textbf{P} & \textbf{R} & \textbf{F$_1$} \\
\midrule
\scriptsize{\textit{(Syntax-aware)}}\\
\citet{lei-etal-2015-high} & n & - & - & 86.6 & - & - & 75.6\\
\citet{fitzgerald-etal-2015-semantic} + Ens  & n & - & - & 86.7 & - & - & 75.2\\
\citet{roth-lapata-2016-neural} + Ens  & n & 90.3 & 85.7 & 87.9 & 79.7 & 73.6 & 76.5\\
\citet{marcheggiani-titov-2017-encoding} + Ens & n & \textbf{90.5} & 87.7 & 89.1 & 80.8 & 77.1 & 78.9 \\ 
\citet{he-etal-2018-syntax} & y & 89.7 & 89.3 & 89.5 & 81.9 & 76.9 & 79.3 \\
\citet{cai-lapata-2019-syntax} + Joint & n & \textbf{90.5} & 88.6 & 89.6 & 80.5 & 78.2 & 79.4\\
\citet{kasai-etal-2019-syntax} & n & 89.0 & 88.2 & 88.6 & 78.0 & 77.2 & 77.6 \\
\citet{he-etal-2019-syntax} & n & 90.0 & \textbf{90.0} & \textbf{90.0} & - & - & - \\
\citet{zhou-etal-2020-parsing} + Joint & n & 88.7 & 89.8 & 89.3 & \textbf{82.5} & \textbf{83.2} & \textbf{82.8} \\
\citet{li-etal-2021-syntax} & n & - & - & 89.2 & - & - & 80.1 \\
\hdashline[1pt/2pt]
\citet{li-etal-2018-unified} + ELMO & n & 90.3 & 89.3 & 89.8 & 80.6 & 79.0 & 79.8\\
\citet{cai-lapata-2019-syntax} + Joint + ELMO & n & 90.9 & 89.1 & 90.0 & 80.8 & 78.6 & 79.7\\
\citet{cai-lapata-2019-semi} + ELMO & n & 91.1 & 90.4 & 90.7 & 82.1 & 81.3 & 81.6\\
\citet{cai-lapata-2019-semi} + ELMO + Semi & n & \textbf{91.7} & 90.8 & 91.2 & 83.2 & 81.9 & 82.5\\
\citet{kasai-etal-2019-syntax} + ELMO & n & 90.3 & 90.0 & 90.2 & 81.0 & 80.5 & 80.8 \\
\citet{he-etal-2019-syntax} + BERT & n & 90.4 & 91.3 & 90.9 & - & - & - \\
\citet{zhou-etal-2020-parsing} + Joint + BERT$_\mathrm{fine-tuned}$ & n & 91.2 & 91.2 & 91.2 & \textbf{85.7} & \textbf{86.1} & \textbf{85.9} \\
\citet{munir2020} + ELMO & n & 91.2 & 90.6 & 90.9 & 83.1 & 82.6 & 82.8\\
\citet{li-etal-2021-syntax} + ELMO & n & 90.5 & \textbf{91.7} & 91.1 & 83.3 & 80.9 & 82.1 \\
\citet{li-etal-2021-syntax} + BERT & n & - & - & \textbf{91.8} & - & - & 83.2 \\
\midrule
\scriptsize{\textit{(Syntax-agnostic)}}\\ 
\citet{marcheggiani-etal-2017-simple} & n & 88.7 & 86.8 & 87.7 & 79.4 & 76.2 & 77.7\\
\citet{cai-etal-2018-full} & y & 89.9 & 89.2 & 89.6 & 79.8 & 78.3 & 79.0\\
\citet{li-etal-2020-high}$^\dagger$ & y & \textbf{91.3} & 88.7 & 90.0 & \textbf{81.8} & 78.4 & 80.0\\
\textbf{This work}$^\dagger$ & y & 90.2 & \textbf{90.5} & \textbf{90.4} & 80.6 & \textbf{80.5} & \textbf{80.6} \\

\hdashline[1pt/2pt]
\citet{li2019dependency}$^\dagger$ + ELMO & n & 89.6 & 91.2 & 90.4 & 81.7 & 81.4 & 81.5 \\
\citet{chen-etal-2019-capturing}$^\dagger$ + ELMO & y & 90.7 & 91.4 & 91.1 & 82.7 & 82.8 & 82.7 \\
\citet{lyu-etal-2019-semantic} + ELMO  & n & - & - & 91.0 & - & - & 82.2 \\
\citet{conia-navigli-2020-bridging}$^\dagger$ + BERT & y & 91.2 & \textbf{91.8} & 91.5 & - & - & 84.6 \\
\citet{li-etal-2020-high}$^\dagger$ + BERT$_\mathrm{fine-tuned}$ & y & \textbf{92.6} & 91.0 & \textbf{91.8} & \textbf{86.5} & 83.8 & \textbf{85.1} \\
\textbf{This work}$^\dagger$ + BERT & y & 91.3 & 91.6 & 91.4 & 84.5 & \textbf{84.2} & 84.4 \\
\bottomrule
\end{tabular}
\centering
\setlength{\abovecaptionskip}{4pt}
\caption{Precision, recall and F$_1$ scores achieved by state-of-the-art SRL systems on the CoNLL-2009 English in-domain (WSJ) and out-of-domain (Brown) test sets \textit{w/ pre-identified predicates}. We use the same 
abbreviations 
%and symbols 
described in Table~\ref{tab:results1} and, additionally, 
\textit{+Semi} indicates semi-supervised training and \textit{+Ens} specifies ensemble system. Please note that, under the \textit{w/ pre-identified predicates} setting, a single-model system only has to perform the three remaining SRL subtasks to be considered an end-to-end approach. Finally, we do not include the syntax-aware transition-based model by \citet{fei-end-SRL}
%in this comparison 
since they do not apply the CoNLL-2009 official scoring script and separately report the accuracy on predicate disambiguation and argument identification+labeling.
}
\label{tab:results2}
\end{table}

\begin{table}[tbp]
\centering
\begin{tabular}{@{\hskip 2pt}l@{\hskip 5pt}c@{\hskip 10pt}c@{\hskip 15pt}c@{\hskip 5pt}c@{\hskip 15pt}c@{\hskip 5pt}c@{\hskip 15pt}c@{\hskip 15pt}c@{\hskip 2pt}}
\toprule
\textbf{System} (\textit{w/ pre-identified predicates}) & \textbf{\scriptsize{end-to-end}} & \textbf{CA$_{id}$} &\textbf{CZ$_{id}$} & \textbf{CZ$_{ood}$} & \textbf{DE$_{id}$} & \textbf{DE$_{ood}$} & \textbf{ES$_{id}$} & \textbf{ZH$_{id}$} \\
\midrule
\scriptsize{\textit{(Syntax-aware)}}\\
CoNLL-2009 ST best & & 80.3 & 86.5 & \textbf{85.4}  &  79.7 & 65.9 & 80.5 & 78.6\\
\citet{zhao-etal-2009-multilingual-dependency} & n & 80.3 & 85.2 & 82.7 & 76.0 & \textbf{67.8} & 80.5 & 77.7  \\
\citet{roth-lapata-2016-neural} + Ens  & n & - & - & - & 80.1 & - & 80.2 & 79.4\\
\citet{cai-lapata-2019-syntax} + Joint & n & - & - & - & 82.7 & - & 81.8 & 83.6\\
\citet{cai-lapata-2019-semi} & n &  - & - & - & 83.3 & - & 82.1 & 84.6\\
\citet{cai-lapata-2019-semi}  + Semi & n &  - & - & - & \textbf{83.8} & - & 82.9 & \textbf{85.0}\\
\citet{he-etal-2019-syntax} & y & \textbf{84.9} & \textbf{88.8} &  - & 78.5 & -  & \textbf{83.9} & 84.8 \\
\hdashline[1pt/2pt]
\citet{he-etal-2019-syntax} + BERT & y & \textbf{86.0} & 89.7 & - & \textbf{81.1} &  - & \textbf{85.2} & \textbf{86.9} \\
\citet{li-etal-2021-syntax} + ELMO & n & 85.5 & \textbf{90.5} & - & 76.6 & - & 84.3 & 86.1 \\
\midrule
\scriptsize{\textit{(Syntax-agnostic)}}\\ 
\citet{marcheggiani-etal-2017-simple} & n & - & 86.0 & 87.2 & - & - & 80.3 & 81.2\\
\citet{mulcaire-etal-2018-polyglot}$^\dagger$ & y & 79.5 & 85.1 & - & 70.0 & - & 77.3 & 81.9 \\
\citet{chen-etal-2019-capturing}  & y & 81.7 & 88.1 & - & 76.4 & - & 81.3 & 81.7 \\
\citet{lyu-etal-2019-semantic}   & n & 80.9 & 87.6 & 86.0 & 75.9 & 65.7 & 80.5 & 83.3 \\
\citet{li-etal-2020-high}$^\dagger$ & y & 84.7 & 90.6 & 90.8 & 76.4 & 70.1 & 84.2 & 86.0\\
\textbf{This work}$^\dagger$  & y & \textbf{85.9} & \textbf{93.3}  & \textbf{92.5} & \textbf{79.5} & \textbf{71.0}  & \textbf{85.2} & \textbf{86.7} \\
\hdashline[1pt/2pt]
\citet{conia-navigli-2020-bridging}$^\dagger$ + BERT & y & 86.2 & 90.1 & 90.6 & 86.5 & 73.1 & 85.3 & 87.3 \\
\citet{li-etal-2020-high}$^\dagger$ + BERT$_\mathrm{fine-tuned}$ & y & \textbf{86.8} & 91.6 & 91.7 & 85.5 & 71.9 & \textbf{86.9} & 88.5 \\
\textbf{This work}$^\dagger$ + BERT  & y & 86.2 & \textbf{93.3} & \textbf{92.2} & \textbf{86.7} & \textbf{75.1}  & 85.7 & \textbf{89.1} \\
\bottomrule
\toprule
\textbf{System} (\textit{w/o pre-identified predicates}) & \textbf{\footnotesize{end-to-end}}  & \textbf{CA$_{id}$} &\textbf{CZ$_{id}$} & \textbf{CZ$_{ood}$} & \textbf{DE$_{id}$} & \textbf{DE$_{ood}$} & \textbf{ES$_{id}$} & \textbf{ZH$_{id}$} \\
\midrule
\scriptsize{\textit{(Syntax-agnostic)}}\\ 
\citet{li-etal-2020-high}$^\dagger$ & n & 83.5  & 89.4 & 89.2  & 60.1 & 43.2 & 83.0 & 81.5 \\
\textbf{This work}$^\dagger$ & y & \textbf{84.6} & \textbf{92.4} & \textbf{91.3} & \textbf{62.0} & \textbf{46.1} & \textbf{84.5} & \textbf{82.6} \\
\hdashline[1pt/1pt]
\citet{li-etal-2020-high}$^\dagger$ + BERT$_\mathrm{fine-tuned}$  & n &  \textbf{85.8} & 90.9 & 90.8 & 67.2 & 41.5 & \textbf{85.8}  & 85.7 \\
\textbf{This work}$^\dagger$ + BERT  & y & 85.2 & \textbf{92.2} & \textbf{91.3} & \textbf{68.5} & \textbf{48.8} & 85.0 & \textbf{86.6}\\
\bottomrule
\end{tabular}
\centering
\setlength{\abovecaptionskip}{4pt}
\caption{F$_1$ scores on the remaining CoNLL-2009 in-domain (${id}$) and out-of-domain (${ood}$) test sets for both \textit{w/ pre-identified predicates} (top) and \textit{w/o pre-identified predicates} (bottom) settings. \textit{CoNLL-2009 ST best} refers to the best F$_1$ scores reported for the CoNLL-2009 shared task \citep{hajic-etal-2009-conll}. We use the same abbreviations 
%and symbols 
previously described in Tables~\ref{tab:results1} and~\ref{tab:results2}. 
}
\label{tab:results3}
\end{table}

\subsection{Results and discussion}
\paragraph{English results} In Table~\ref{tab:results1}, we compare our model against previous full SRL systems on English in-domain and out-of-domain tests sets under the \textit{w/o pre-identified predicates} setting. Our single-model approach achieves the best performance on both test sets among syntax-agnostic SRL systems without deep contextualized word embeddings, even outperforming all syntax-based models on the WSJ test set. When pre-trained language models come into play, our system obtains competitive accuracies (improving over again all syntax-aware approaches on the in-domain test set), but it is slightly surpassed by those graph-based models \citep{li-etal-2020-high,zhou2021fast} that adapt BERT-based embeddings to SRL by fine-tuning them during training.

Table~\ref{tab:results2} presents the results on English in-domain and out-of-domain tests sets \textit{w/ pre-identified predicates}. It can be seen as predicate-centered SRL systems outnumber those developed for full SRL (which are reported in Table~\ref{tab:results1}). While our approach was originally designed for 
dealing with the lack of gold predicates and just minor adaptations were undertaken for leveraging pre-identified predicate information, our system behaves similarly to the full SRL setting: it is the best-performing syntax-agnostic approach without contextualized word representations (also surpassing syntax-aware systems on the WSJ test set), but the highest scores are reported by graph-based models that fine-tune BERT-based embeddings \citep{li-etal-2020-high} or, while keeping them frozen, leverage syntactic information \citep{li-etal-2021-syntax}. 

\paragraph{Multilingual results} Table~\ref{tab:results3} summarizes the performance on the remaining CoNLL-2009 languages (including out-of-domain test sets if available) under both \textit{w/ pre-identified predicates} and \textit{w/o pre-identified predicates} settings. Compared with previous methods, our approach yields strong performance consistent across languages, regardless of the availability of gold predicate information. 

Without contextualized word representations, our end-to-end SRL system improves over the best 
%previous 
syntax-agnostic methods on all languages in the in-domain setting (with and without pre-identified predicates), bringing notable improvements on both high-resource (\textit{e.g.}, Czech) and low-resource (\textit{e.g.}, German) datasets. Our proposal also outperforms syntax-based models in all datasets except German, meaning that leveraging syntactic information \citep{cai-lapata-2019-semi} is still crucial for obtaining state-of-the-art results on that language.
%in-domain test set.% without BERT-based embeddings.

When our model is augmented with frozen BERT-based embeddings, we achieve the highest score to date on 3 out of 5 languages in the in-domain setting (with and without given predicates), being only outperformed by \citep{li-etal-2020-high} on Catalan and Spanish, where fine-tuning contextualized word vectors yields substantial accuracy gains in these low-resource datasets.

Regarding 
%and proving its robustness %of our approach is also evident when considering results 
results on the out-of-domain data (with and without gold predicates), our model (with and without BERT-based embeddings) 
%proves its robustness by 
%obtaining a notable performance 
%surpassing 
outperforms existing SRL systems (including syntax-aware approaches) by a wide margin on Czech and German test sets (being especially challenging the latter, since it contains numerous infrequent predicates specifically included for the CoNLL-2009 shared task). Lastly, 
%similarly to \citep{li-etal-2020-high}, 
we observe in Czech as adding deep contextualized word embeddings has no effect without given predicates and is harmful with pre-identified predicates, probably meaning that a task-specific fine-tuning would be helpful in that case.
%A transition-based approach proves to be more effective in multilingual SRL, especially in low-resource languages such Catalan, German and Spanish.

Finally, it is worth mentioning that, to the best of our knowledge, our proposal is the first end-to-end system that provides scores for full multilingual SRL (without given predicates) on CoNLL-2009 datasets, since \citep{li-etal-2020-high} (the only graph-based model included in that setting) adopts a pipeline strategy. Moreover, our model is the best-performing approach
%on all languages tested 
among truly syntax-agnostic SRL systems, 
%which use neither PoS tag nor deep contextualized word embeddings. 
\change{which do not exploit PoS tag embeddings.}
Lastly, we do not fine-tune hyperparameters for individual languages, suggesting that
the presented approach is robust and can be directly applied to other languages.

\subsection{Ablation study}
The previous section has already shown as BERT-based embeddings substantially boost our model accuracy; however, we do not know the performance impact of, for instance, beam-search decoding or high-order features provided by
%used when we leverage the encoder representation of 
the last assigned predicate. Thus, we conduct an ablation study of our neural architecture in order to better understand the contribution of each component in the final accuracy. In particular, we successively remove from the full model:  
the beam-search decoding, co-parent features (\textit{i.e.}, state configuration representations $s_t$ are generated without the addition of $h_j$ to $h_i$), lemma embeddings and character embeddings. In Table~\ref{tab:study}, we can observe that the removal of every component leads to an overall performance degradation; however, %these results show that 
%being 
character embedding ablation has the largest impact on the performance of our model, resulting in significant drops in F$_1$ ($-1.68$), F$^{pred}_1$ ($-1.73$) and F$^{arg}_1$ scores ($-1.68$).
 %proving to have a notable contribution to the performance of our model. 
 Lemma embeddings also play an important role in our neural architecture, notably increasing our model performance. In addition, we can also see as the lack of co-parent features improves the accuracy on predicate identification and disambiguation (F$^{pred}_1$), but penalizes the performance on argument identification and labeling (F$^{arg}_1$). This means that co-parent features are especially beneficial for argument processing subtasks, which are more complex and have a larger impact on the overall F$_1$ score (since, except in Czech, there are significantly more arguments than predicates in CoNLL-2009 corpora). Finally, the beam-search decoding with beam size 5 has a minor impact in comparison to the addition of lemma and character embeddings. We think that a further beam-size exploration might probably increase its contribution to the final accuracy.

%have a notable contribution in our model's performance.  

%In-depth analysis is con- ducted to uncover the important components of our final model, which can help comprehensive understanding of our model.

\begin{table}[tbp]
\centering
\begin{tabular}{@{\hskip 2pt}lc@{\hskip 5pt}c@{\hskip 5pt}l@{\hskip 5pt}l@{\hskip 5pt}l@{\hskip 2pt}}
\toprule
\textbf{System} & \textbf{P} & \textbf{R} & \textbf{\ \ F$_1$} & \textbf{\ \ F$^{pred}_1$} & \textbf{\ \ F$^{arg}_1$}\\
\midrule
\textbf{Full model} & 85.36 & 85.89 & 85.63 & 90.34 & 83.46 \\
\ \ - beam search & 84.69 & 86.22 & 85.44$^{\mathrm{(-0.19)}}$ & 90.20$^{\mathrm{(-0.14)}}$ & 83.26$^{\mathrm{(-0.20)}}$\\
\ \ \ - co-parent features & 84.47 & 86.13 & 85.29$^{\mathrm{(-0.15)}}$ & 90.45$^{\mathrm{(+0.25)}}$ & 82.91$^{\mathrm{(-0.35)}}$\\
\ \ \ \ - lemma embeddings & 82.65 & 86.23 & 84.40$^{\mathrm{(-0.89)}}$ & 89.77$^{\mathrm{(-0.68)}}$ & 81.96$^{\mathrm{(-0.95)}}$ \\
\ \ \ \ \  - character embeddings & 80.94 & 84.58 & 82.72$^{\mathrm{(-1.68)}}$ & 88.04$^{\mathrm{(-1.73)}}$ & 80.28$^{\mathrm{(-1.68)}}$\\
\bottomrule
\end{tabular}
\centering
\setlength{\abovecaptionskip}{4pt}
\caption{Overall precision, recall and F$_1$ scores, as well as specific F$_1$ scores measured only on predicate identification+disambiguation (F$^{pred}_1$) and argument identification+labeling (F$^{arg}_1$) subtasks 
%of the ablation study
on the CoNLL-2009 English development split under the \textit{w/o pre-identified predicates} setup.}
\label{tab:study}
\end{table}

\subsection{Time complexity}
\label{sec:complexity}
The full time complexity of best-performing graph-based models \citep{li-etal-2020-high,zhou2021fast} is $O(n^3)$ due to the leverage of higher-order information. We will prove that our approach is more efficient, being $O(n^2)$ its overall expected worst-case
running time for the range of data tested in our experiments. 

\begin{figure}
\centering
\includegraphics[width=0.33\textwidth]{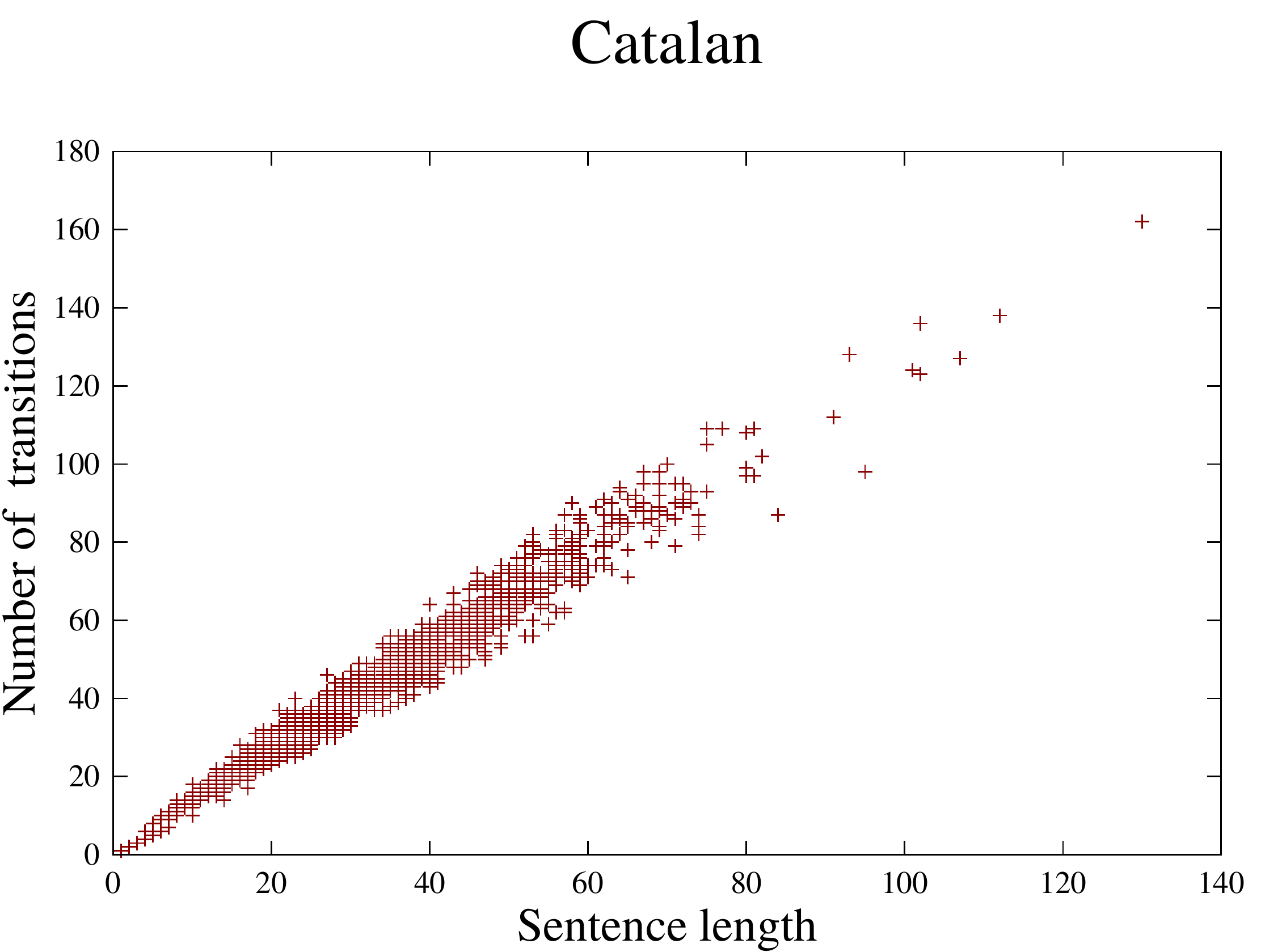}\includegraphics[width=0.33\textwidth]{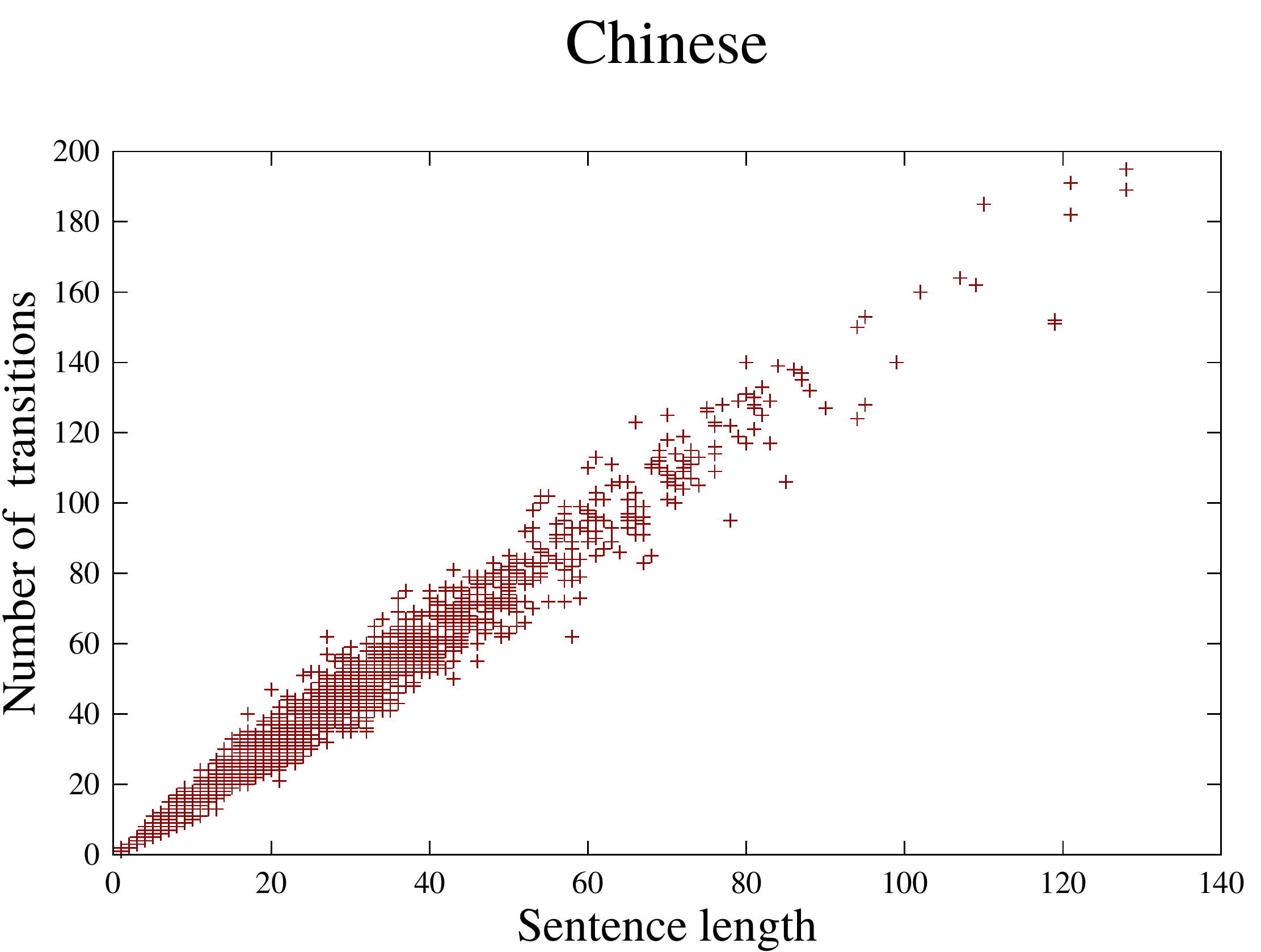}\includegraphics[width=0.33\textwidth]{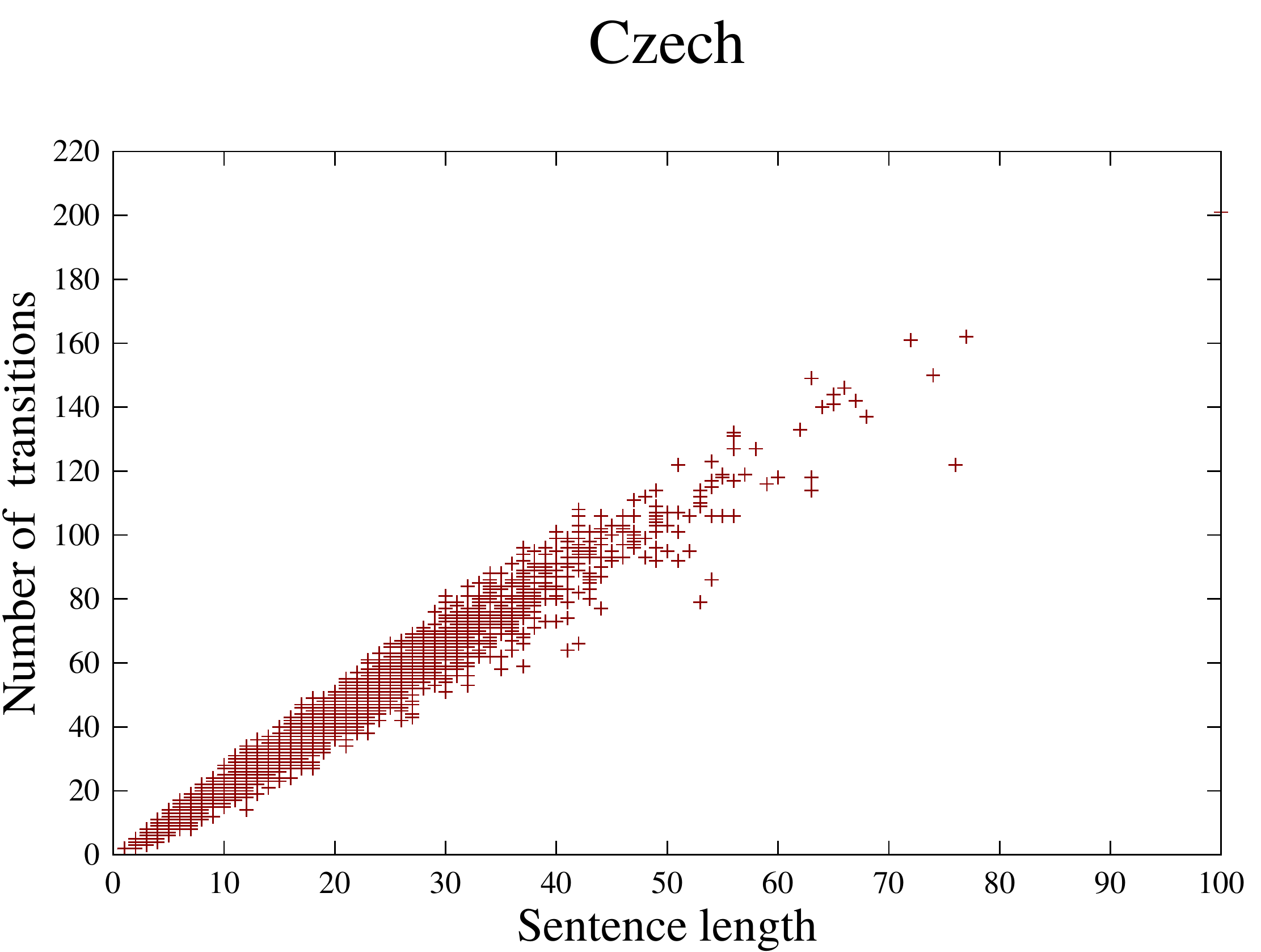} \\
\vspace{5pt}
\includegraphics[width=0.33\textwidth]{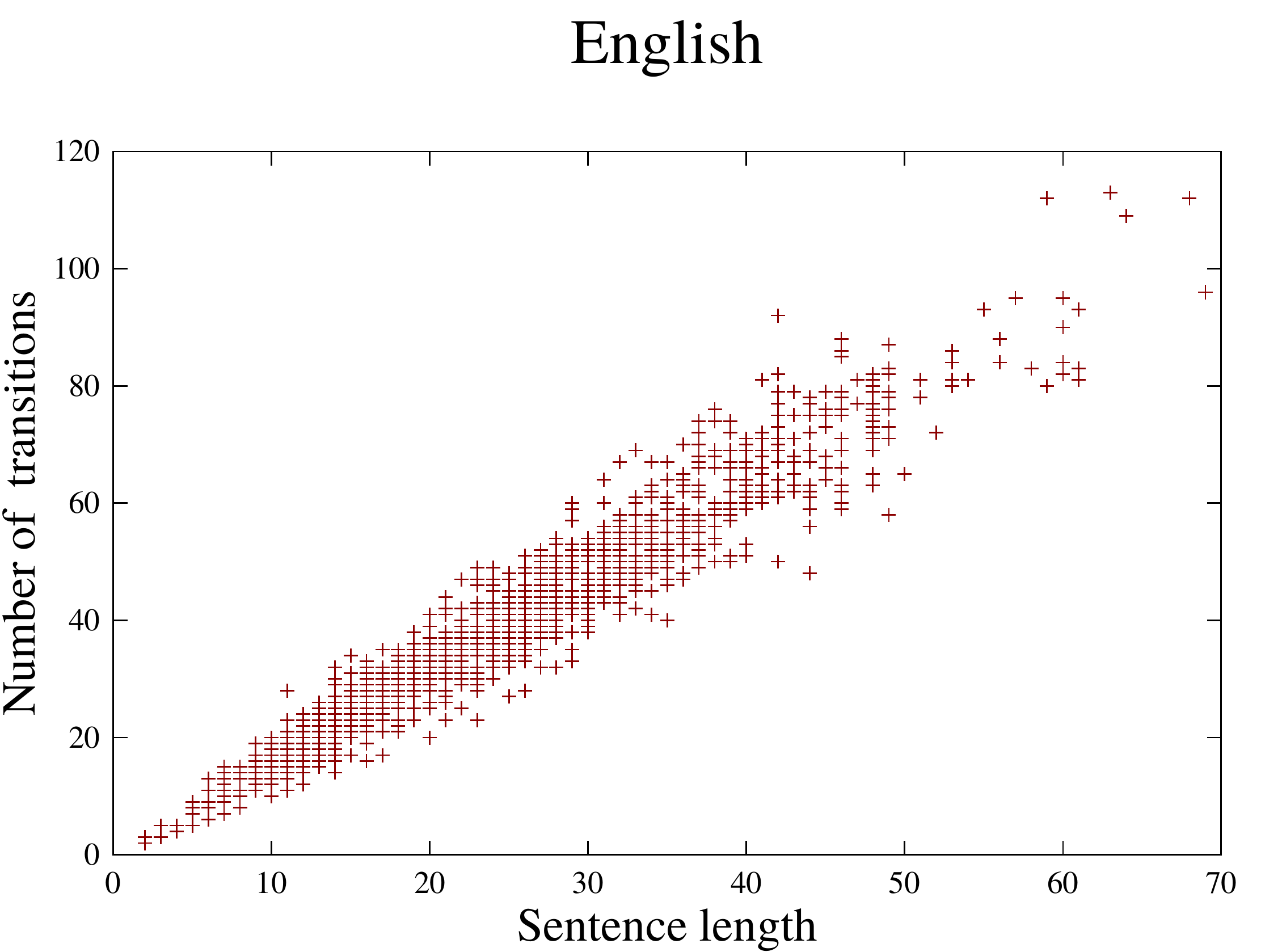}\includegraphics[width=0.33\textwidth]{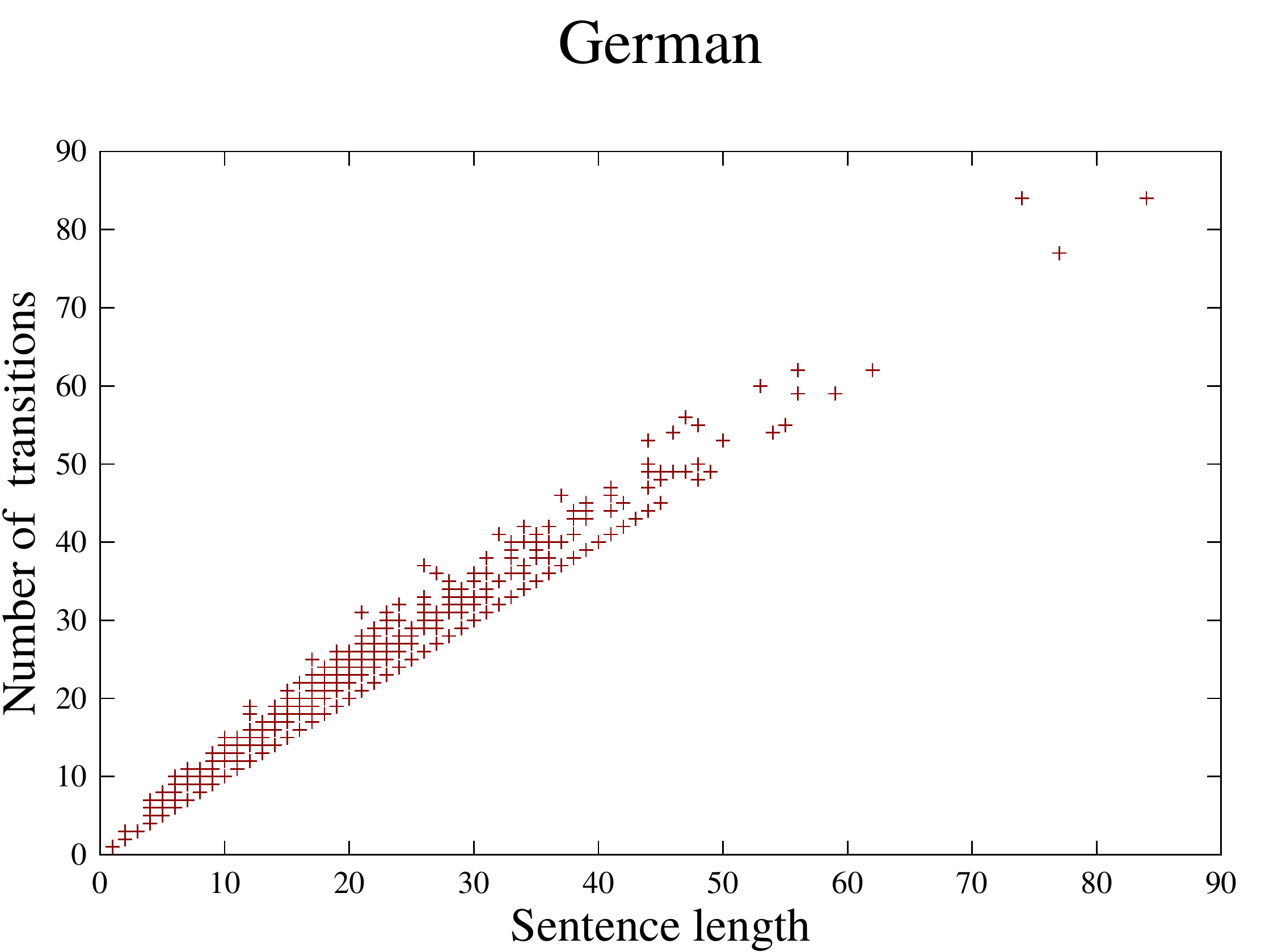}\includegraphics[width=0.33\textwidth]{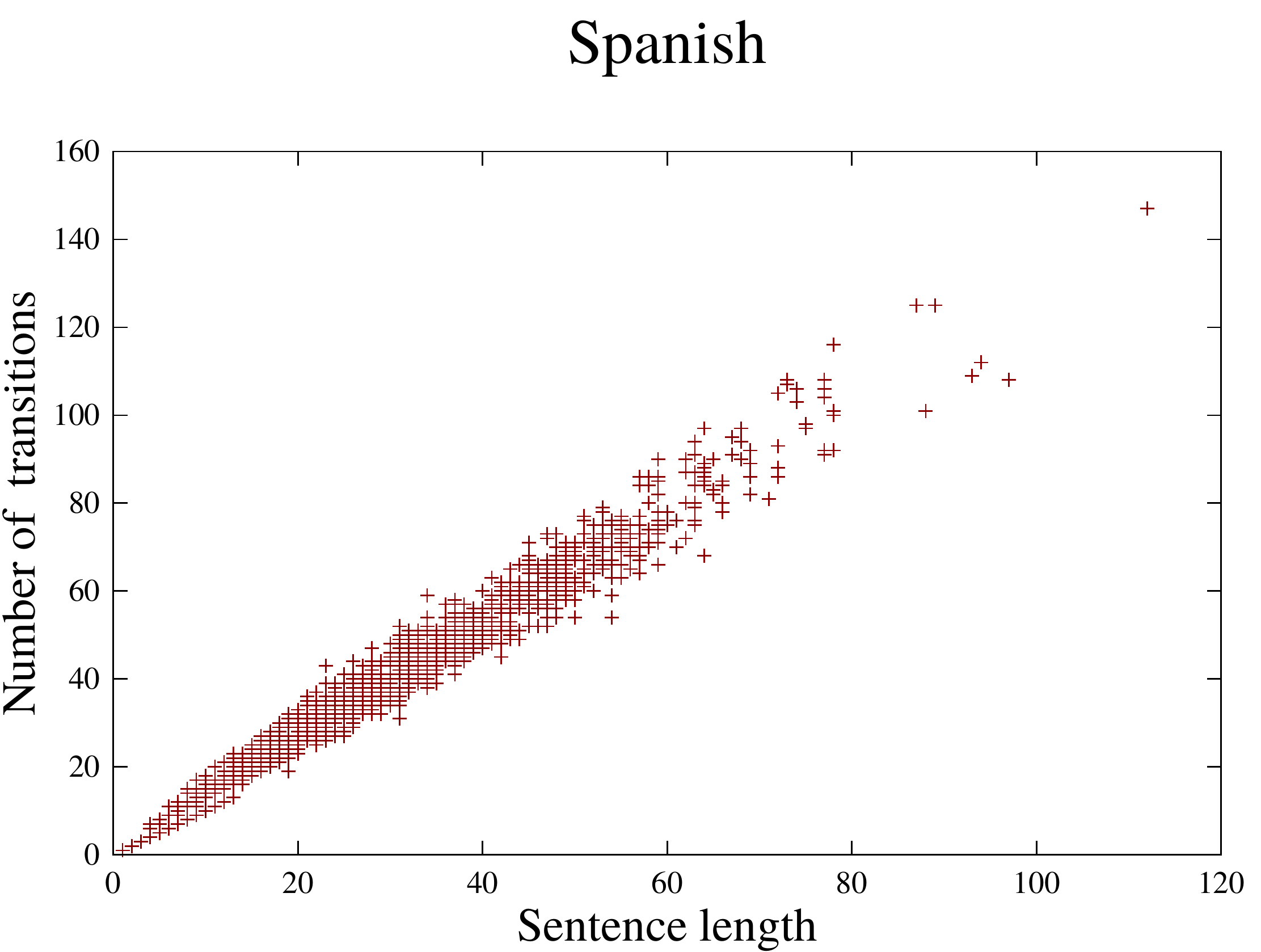}
\caption{Number of transitions predicted by our model relative to the sentence length for CoNLL-2009 development sets.}
\label{fig:complexity}
\end{figure}

Being $n$ the sentence length, a general directed graph can have at most $\Theta(n^2)$ edges, requiring our transition system $O(n^2)$ actions to build it in the worst case (\textit{i.e.}, $n$ \textsc{Shift} transitions for processing all words and $n$ \textsc{Arc} actions per word for assigning all its heads). Nevertheless, predicate-argument graphs from CoNLL-2009 datasets can be produced with $O(n)$ transitions. In order to prove that, we need to determine the 
%parsing 
complexity of the proposed transition system in practice. This can be done by examining, for each sentence, 
how the predicted transition sequence length varies as a function of sentence length
%the relation between the transition sequence length predicted by our model 
%with respect to 
%and 
%the sentence length 
\citep{Kubler2009}. 
Figure~\ref{fig:complexity} graphically shows the relation between the number of predicted transitions and the number of words for every sentence from CoNLL-2009 development splits. We can clearly observe a linear relationship across all languages, which means that the number of \textsc{Arc} transitions required per word is notably low and behaves 
like a constant in the represented linear function. This behavior is supported  
%behind this 
by the fact that, due to the significant amount of unattached words, there are substantially less predicate-argument edges than words in graphs 
%generated 
from 
%SRL structures in 
CoNLL-2009 data,
%in the sentence, 
being the average ratio of edges per word in a sentence less than 1 in practically all training sets: 0.32 in Catalan, 0.53 in Chinese, 0.08 in German, 0.56 in English and 0.34 in Spanish. The exception is observed in graphs from the Czech dataset, where we have more than one edge per word on average (1.15). From this information, we can state that every sentence from CoNLL-2009 corpora (except Czech)
%\footnote{Since some graphs from Czech can have more than one edge per word on average, } 
can be processed with 2$n$ transitions at most  (\textit{i.e.}, $n$ \textsc{Shift} actions plus $n$ \textsc{Arc} transitions); and we will require 3$n$ transitions in the worst case for generating graphs from Czech (\textit{i.e.}, $n$ \textsc{Shift} transitions plus 2$n$ \textsc{Arc} actions).
In both cases, the resulting number of transitions is linear.

Finally, the time complexity of our approach not only comes from the transition system: for predicting each transition, the attention vector $\alpha_t$ must be computed over the whole input sentence in $O(n)$ time. Consequently, the overall time complexity of the proposed SRL system on CoNLL-2009 corpora is $O(n^2)$.

\section{Conclusions}
\label{sec:conclusion}

In this article, we propose the first syntax-agnostic transition-based approach for full end-to-end SRL. This exclusively relies on raw text as input, neither requiring syntactic trees nor resorting to external models to accomplish any of the SRL subtasks. In addition, we prove that our technique is more efficient in practice ($O(n^2)$) than best-performing graph-based models ($O(n^3)$). Thanks to these advantageous features, it can be easily applied in real-world applications and low-resource languages, where, for instance, syntactic information is scarce.

 We not only extensively evaluate our model on in-domain and out-of-domain CoNLL-2009 corpora under the full SRL setting, but additionally adapt our approach to handle gold predicate information in order to perform a fair comparison against the vast majority of previous methods, which do not address predicate identification. While our single-model proposal obtains competitive accuracies on the CoNLL-2009 English data, it excels in the remaining five languages, achieving a strong performance across in-domain and out-of-domain test sets as well as high-resource and low-resource languages. 
 
 Lastly, although   
%Regardless of gold predicate information, 
our transition-based SRL system is robust and accurate, it is outperformed by graph-based models that either fine-tune deep contextualized word embeddings or use additional syntactic information. 
%Even following a cost-effective strategy when deep contextualized word embeddings augment our neural model, we manage to provide competitive accuracies in comparison to those best-performing systems that fine-tune them.
%Finally, it is worth mentioning that
Therefore, while we think that leveraging syntactic trees makes SRL systems less cost-effective and more dependent on high-resource languages, 
%we do not have any bias to syntax-agnostic approaches and 
our model can 
%certainly 
benefit from syntax to further improve its performance. And, additionally, we could also perform a task-specific fine-tuning of BERT-based embeddings to obtain substantial accuracy gains in low-resource languages. 
%can be also applied to boost the model performance.

\section*{Acknowledgments}
We acknowledge the European Research Council (ERC), which has funded this research under the European Union’s Horizon 2020 research and innovation programme (FASTPARSE, grant agreement No 714150), ERDF/MICINN-AEI (SCANNER-UDC, PID2020-113230RB-C21), Xunta de Galicia (ED431C 2020/11), and Centro de Investigaci\'on de Galicia ``CITIC'', funded by Xunta de Galicia and the European Union (ERDF - Galicia 2014-2020 Program), by grant ED431G 2019/01.
%Funding for open access charge: Universidade da Coruña/CISUG.

\printcredits

\bibliographystyle{cas-model2-names}

\bibliography{anthology,main,bibliography}

% \appendix

\end{document}